\documentclass{article} 
\usepackage{iclr2019_conference,times}
\usepackage{lipsum}

\usepackage{amsmath,amsfonts,bm}

\def\eqref#1{equation~\ref{#1}}

\def\1{\bm{1}}

\DeclareMathAlphabet{\mathsfit}{\encodingdefault}{\sfdefault}{m}{sl}
\SetMathAlphabet{\mathsfit}{bold}{\encodingdefault}{\sfdefault}{bx}{n}

\DeclareMathOperator*{\argmax}{arg\,max}

\usepackage{sidecap}
\usepackage{hyperref}
\usepackage{url}
\usepackage{setspace}
\usepackage{bm}
\usepackage{tikz}
\usepackage{caption}
\usepackage{subcaption}
\usepackage{bm}
\usepackage[most]{tcolorbox}
\usepackage{mathtools}
\tcbset{colback=green!15!white, colframe=black, 
        highlight math style= {enhanced, 
            colframe=red,colback=red!10!white,boxsep=0pt}
        }

\makeatletter
\newcommand{\printfnsymbol}[1]{%
  \textsuperscript{\@fnsymbol{#1}}%
}
\makeatother

\newcommand\blfootnote[1]{%
  \begingroup
  \renewcommand\thefootnote{}\footnote{#1}%
  \addtocounter{footnote}{-1}%
  \endgroup
}

\title{PIE: Pseudo-Invertible Encoder}

\author{Jan Jetze Beitler\thanks{Equal contribution} \\
Institute of Informatics\\
University of Amsterdam\\
Netherlands\\
\And
Ivan Sosnovik\printfnsymbol{1} \& Arnold Smeulders \\
UvA-Bosch Delta Lab \\
University of Amsterdam\\
Netherlands
}

\input{equations}
\usetikzlibrary{intersections}
\usetikzlibrary{fit}
\usetikzlibrary{calc}
\usetikzlibrary{backgrounds, positioning, patterns, decorations.pathreplacing}
\usetikzlibrary{intersections}
\usetikzlibrary{fit}				
\usetikzlibrary{backgrounds, positioning}
\usetikzlibrary{patterns}

\definecolor{pool color}{RGB}{215,25,28}
\definecolor{split color}{RGB}{253,174,97}
\definecolor{coupling color}{RGB}{255,255,191}
\definecolor{affine color}{RGB}{171,221,164}
\definecolor{norm color}{RGB}{43,131,186}

\tikzstyle{func}=[circle, thick, minimum size=0.5cm, draw=blue!80, fill=blue!20]
\tikzstyle{operator}=[circle, thick, minimum size=0.5cm, draw=black]
\tikzstyle{apply}=[rectangle, rounded corners, thin, minimum size=1.0cm, draw=blue!20, fill=blue!20]
\tikzstyle{background}=[rectangle, inner sep=0.2cm, rounded corners=5mm]
\tikzstyle{support}=[coordinate]

\tikzstyle{layer}=[draw=black,very thick, minimum width=20pt, minimum height=30pt, rounded corners=1mm]
\tikzstyle{block}=[draw=black, very thick, minimum width=60 pt, minimum height=40pt]
\tikzstyle{affine}=[fill=affine color, layer]
\tikzstyle{batch_norm}=[fill=norm color, layer]
\tikzstyle{pooling}=[fill=pool color, layer]
\tikzstyle{coupling}=[fill=coupling color, layer]
\tikzstyle{split}=[fill=split color, layer]
\tikzstyle{contour}=[very thick, rectangle, fill=none, draw=black, dashed, inner sep=8pt, rounded corners=10pt]

\tikzstyle{arrow}=[->, line width=1pt]
\iclrfinalcopy 
\begin{document}

\maketitle
\begin{abstract}
We consider the problem of information compression from high dimensional data. 
Where many studies consider the problem of compression by non-invertible transformations, 
we emphasize the importance of invertible compression. We introduce a new class of likelihood-based autoencoders with pseudo bijective architecture, which we call Pseudo Invertible Encoders. We provide the theoretical explanation of their principles. We evaluate Gaussian Pseudo Invertible Encoder on MNIST, where our model outperforms WAE and VAE in sharpness of the generated images.
\blfootnote{Correspondence to Ivan Sosnovik: \texttt{i.sosnovik@uva.nl}}
\end{abstract}

\section{Introduction}

We consider the problem of information compression from high-dimensional data. Where many studies consider the problem of compression by non-invertible transformations, we emphasize the importance of invertible compression as there are many cases where one cannot or will not decide \textit{a priori} what part of the information is important and what part is not. Compression of images for person ID in a small company requires less resolution than person ID at an airport. To lose a part of the information without harm to the future purpose of viewing the picture requires knowing the purpose upfront. Therefore, the fundamental advantage of invertible information compression is that compression can be undone if a future purpose requires so. 

Recent advances in classification models have demonstrated that deep learning architectures of proper design do not lead to information loss while still being able to achieve state-of-the-art in classification performance. These \textit{i}-RevNet models \citep{iRevNet} implement a small but essential modification of the popular ResNet models while achieving invertibility and a performance similar to the standard ResNet \citep{He2016}. This is of great interest as it contradicts the intuition that information loss is essential to achieve good performance in classification \citep{tishby2015deep}. Despite the requirement of the invertibility, flow-based generative models \citep{NICE,RealNVP,Rezende2015,Glow} demonstrate that the combination of bijective mappings allows one to transform the raw distribution of the input data to any desired distribution and perform the manipulation of the data. 

On the other hand, autoencoders have provided the ideal mechanism to reduce the data to the bare minimum while retaining all essential information for a specific task, the one implemented in the loss function. Variational autoencoders (VAE) \citep{Kingma2013} and Wasserstein autoencoders (WAE) \citep{Tolstikhin2018} are performing best. They provide an approach for stable training of autoencoders, which demonstrates good results at reconstruction and generation. However, both of these methods involve the optimization of the objective defined on the pixel level. We would emphasize the importance of avoiding the separate decoder part and training the model without relying on the reconstruction quality directly.

\begin{figure}
    \centering
    \includegraphics[width=0.7\linewidth]{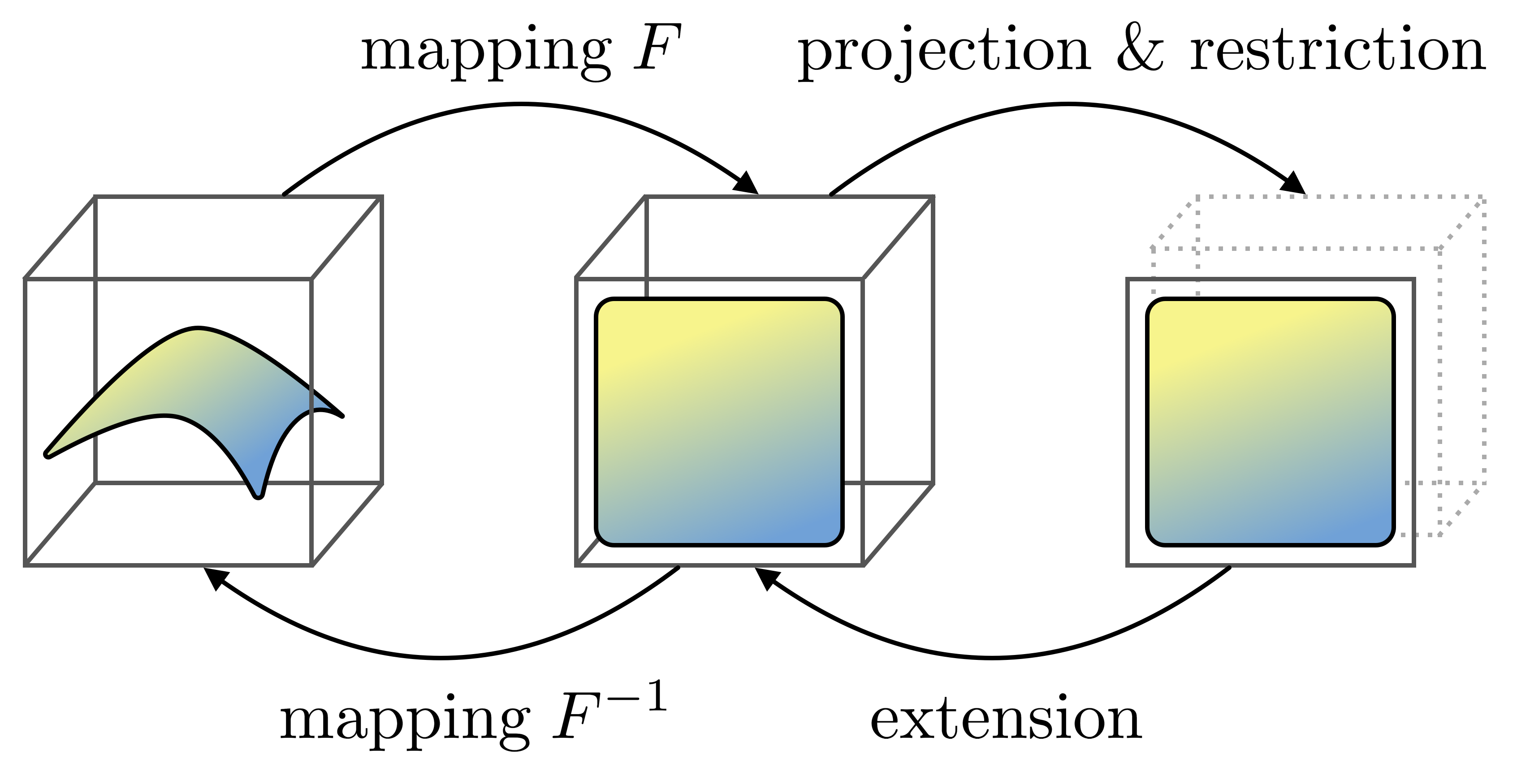}
    \caption{Schematic representation of the proposed mechanism for dimensionality reduction.}
    \label{scheme:main}
\end{figure}

Combining the best of invertible mappings and autoencoders, we introduce Pseudo Invertible Encoder. Our model combines bijective functions with restriction and extension of the mappings to the dependent sub-manifolds Fig. \ref{scheme:main}. 
The main contributions of this paper are the following:
\begin{itemize}
    \item We introduce a new class of likelihood-based autoencoders, which we call Pseudo Invertible Encoders. We provide the theoretical explanation of their principles.
    \item We demonstrate the properties of Gaussian Pseudo Invertible Encoder in manifold learning.
    \item We compare our model with WAE and VAE on MNIST, and report that the sharpness of the images, generated by our models is better.
\end{itemize}

\section{Related Work}
\subsection{Invertible models}
ResNets \citep{He2016} allow for arbitrary deep networks and thus memory consumption becomes a bottleneck. \citep{RevNet} propose a Reversible Residual Network (RevNet) where each layer's activations can be reconstructed from the activations of the next layer. By replacing the residual blocks with coupling layers, they mimic the behavior of residual blocks while being able to retrieve the original input of the layer. RevNet replaces the residual blocks of ResNets but also accommodates non-invertible components to train more efficiently. By adding a downsampling operator to the coupling layer, $i$-RevNet circumvents these non-invertible modules \citep{iRevNet}. With this, they show that losing information is not a necessary condition to learn representations that generalize well on complicated problems. Although $i$-RevNet circumvents non-invertible modules, data is not compressed and the model is only invertible up to the last layer. These methods do not allow for dimensionality reduction. In the current research, we build a pseudo invertible model which performs dimensionality reduction.
    
\subsection{Autoencoders}
Autoencoders were first introduced by \citep{Rummelhart1986} as an unsupervised learning algorithm. They are now widely used as a technique for dimensionality reduction by compressing input data. By training an encoder and a decoder network, and measuring the distance between the original and the reconstructed data, data can be represented in a latent space. The latent codes can then be used for supervised learning algorithms. Instead of learning a compressed representation of the input data \citep{Kingma2013} propose to learn the parameters of a probability distribution that represents the data. \citep{tolstikhin2017wasserstein} introduced a new class of models --- Wasserstein Autoencoders, which use Optimal Transport to be trained. These methods require the optimization of the objective function which includes the terms defined on the pixel level. Our model does not require such optimization. Moreover, it only performs encoding at training time.

\section{Theory}
\begin{figure}
    \centering
    \includegraphics[width=0.6\linewidth]{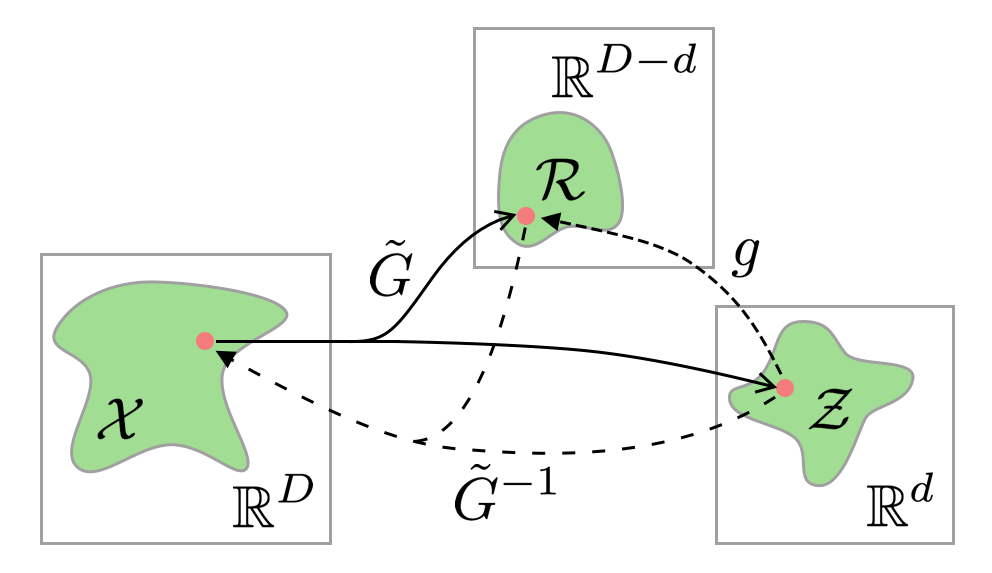}
    \caption{The schematic representation of the \textit{Restriction-Extension} approach.
    The invertible mapping $\mathcal{X} \leftrightarrow \mathcal{Z}$ 
    is preformed by using the dependent sub-manifold $\mathcal{R} = g(\mathcal{Z})$ 
    and a pair of extended functions $\tilde{G},\; \tilde{G}^{-1}$.}
    \label{scheme:manifolds}
\end{figure}

Here we introduce the approach for performing dimensionality reduction with invertible mappings. Our method is based on the \textit{restriction} of the mappings to low-dimensional manifolds, and the \textit{extension} of the inverse mappings with certain constraints (Fig. \ref{scheme:manifolds}).

\subsection{Restriction-Extension Approach} 

Given a data $\mathbf{x}_i \in \mathcal{X} \subset \mathbb{R}^D$. 
Assuming that $\mathcal{X}$ is a $d$-dimensional manifold, 
with $d < D$, we seek to find a mapping $G: \mathbb{R}^D \rightarrow \mathbb{R}^d$ invertible on $\mathcal{X}$. 
In other words, we are looking for a pair of associated functions $G$ and $G^{-1}$ such that 
\begin{equation}
    \label{eq:initial}
    \begin{cases}
        G(\mathcal{X}) = \mathcal{Z} \subset \mathbb{R}^d \\
        G^{-1}(\mathcal{Z}) = \mathcal{X} 
    \end{cases}
\end{equation}

Let $\mathcal{R}$ be an open set in $\mathbb{R}^{D-d}$. 
We use this residual manifold in order to match the dimensionalities 
of the hidden and the initial spaces.
Here we introduce a function $g: \mathbb{R}^d \rightarrow \mathbb{R}^{D-d}$.
With no loss of generality we can say that $\mathcal{R} = g(\mathcal{Z})$. 
We use the pair of extended functions $\tilde{G}: \mathbb{R}^D \rightarrow \mathbb{R}^d \times \mathbb{R}^{D-d}$ and $\tilde{G}^{-1}: \mathbb{R}^d \times \mathbb{R}^{D-d} \rightarrow \mathbb{R}^D$ to rewrite Eq. \ref{eq:initial}:
\begin{equation}
    \label{eq:extended}
    \begin{cases}
      \tilde{G}(\mathcal{X}) = \mathcal{Z} \times \mathcal{R}  \\
      \tilde{G}^{-1}(\mathcal{Z} \times \mathcal{R}) = \mathcal{X}
    \end{cases}
\end{equation}

Rather than searching for the invertible dimensionality reduction mapping directly, 
we seek to find $\tilde{G}$, an invertible transformation with certain constraints, 
expressed by $\mathcal{R}$.

In search for $\tilde{G}$, we focus on $F_{\bm{\theta}}: \mathbb{R}^D \rightarrow \mathbb{R}^D$, 
$F_{\bm{\theta}} \in \mathcal{F}$, where $\mathcal{F}$ is a parametric family of 
functions invertible on $\mathbb{R}^D$.  
We select a function $F_{\bm{\theta}}$ with parameters $\bm{\theta}$ which satisfies the constraint:
\begin{equation}
  \label{eq:proj_constraint}
    F_{\bm{\theta}}^{-1} \circ P_{\mathbb{R}^d \times \mathcal{R}} \circ F_{\bm{\theta}} 
    = \text{id}_\mathcal{X}
\end{equation}
where $P_{\mathbb{R}^d \times \mathcal{R}}$ is the orthogonal projection from $\mathbb{R}^d \times \mathbb{R}^{D-d}$
to $\mathbb{R}^d \times \mathcal{R}$. 

Taking into account constraint \ref{eq:proj_constraint}, we derive 
$F_{\bm{\theta}}(\mathbf{x}) = [\mathbf{z}, \mathbf{r}]$, where $\mathbf{z} \in \mathcal{Z}$ and 
$\mathbf{r} \in \mathcal{R}$.
By combining this with Eq. \ref{eq:extended} we have the desired pair of functions: 
\begin{equation}
  \label{eq:final}
  \begin{cases}
    G(\mathbf{x})= \mathbf{z},\\
    G^{-1}(\mathbf{z}) = F_{\bm{\theta}}^{-1}([\mathbf{z}, g(\mathbf{z})])
  \end{cases}
\end{equation}

The obtained function $G$ is \textit{Pseudo Invertible Endocer}, or shortly \textit{PIE}. 

\subsection{Log Likelihood Maximization}
As we are interested in high dimensional data such as images, 
the explicit choice of parameters $\bm{\theta}$ is impossible. 
We choose $\bm{\theta}^*$ as a maximizer of the log likelihood of the observed data given the prior $p_\theta(\mathbf{x})$:

\begin{equation}
  \bm{\theta}^* = \argmax_{\bm{\theta}} [\log p_{\bm{\theta}}(\mathbf{x})]
\end{equation}

After a change of variables according to Eq. \ref{eq:final} we obtain
\begin{equation}
  p(\mathbf{x}) = p(F_{\bm{\theta}}(\mathbf{x})) \Big|\det 
  \Big(\frac{\partial F_{\bm{\theta}}}{\partial \mathbf{x}^T}\Big)\Big|
\end{equation}
Taking into account the constraint \ref{eq:proj_constraint} we derive the joint distribution 
for $F_{\bm{\theta}}(\mathbf{x}) = [\mathbf{z}, \mathbf{r}]$
\begin{equation}
  p(F_{\bm{\theta}}(\mathbf{x})) = p(\mathbf{z}, \mathbf{r}) = p(\mathbf{r}| \mathbf{z}) p(\mathbf{z})
\end{equation}

\begin{equation}
    \int\displaylimits_{\mathcal{X}}p(F_{\bm{\theta}}(\mathbf{x})) d \mathbf{x} =
    \int\displaylimits_{\mathcal{R} = g(\mathcal{Z})}
    \int\displaylimits_{\mathcal{Z}} p(\mathbf{r}| \mathbf{z}) p(\mathbf{z}) d \mathbf{r} d \mathbf{z}=
    \int\displaylimits_{\mathcal{R}}
    \int\displaylimits_{\mathcal{Z}} \delta(\mathbf{r} - g(\mathbf{z})) p(\mathbf{z}) d \mathbf{r} d \mathbf{z}
\end{equation}

\begin{equation}
  p(F_{\bm{\theta}}(\mathbf{x}))= \delta(\mathbf{r} - g(\mathbf{z})) p(\mathbf{z})
\end{equation}

Dirac's delta function can be viewed as a limit of a sequence of Gaussians: 
\begin{align}
    \label{eq:delta}
  \delta (\mathbf{x}) = \lim_{\epsilon \rightarrow 0} \mathcal{N}(\mathbf{x}| \bm{0}, \epsilon^2\mathbf{I})
\end{align}

Let us fix $\epsilon^2 = \epsilon_0^2 \ll 1$. Then 
\begin{align}
  & \delta (\mathbf{x}) \approx \mathcal{N}(\mathbf{x}| \bm{0}, \epsilon_0^2 \mathbf{I})\\
  & \delta (\mathbf{r} - g(\mathbf{z})) \approx \mathcal{N}(\mathbf{r}| g(\mathbf{z}), \epsilon_0^2 \mathbf{I})
\end{align}

Finally, for the log likelihood we have:
\begin{equation}
  \tcboxmath{\log p(\mathbf{x}) \approx \log p(\mathbf{z}) + 
  \log \mathcal{N}(\mathbf{r} | g(\mathbf{z}), \epsilon_0^2 \mathbf{I}) +
  \log \Big|\det \Big(\frac{\partial F_{\bm{\theta}}}{\partial \mathbf{x}^T}\Big)\Big| }
\end{equation}

We choose a prior distribution $p(\mathbf{z})$ as Standard Gaussian.
We search for the parameters $\pmb{\theta}^*$ by using Stochastic Gradient Descent.

\subsection{Composition of Bijectives} 

\begin{figure}
  \begin{subfigure}{.33\textwidth}
    \centering
    \includegraphics[width=0.95\linewidth]{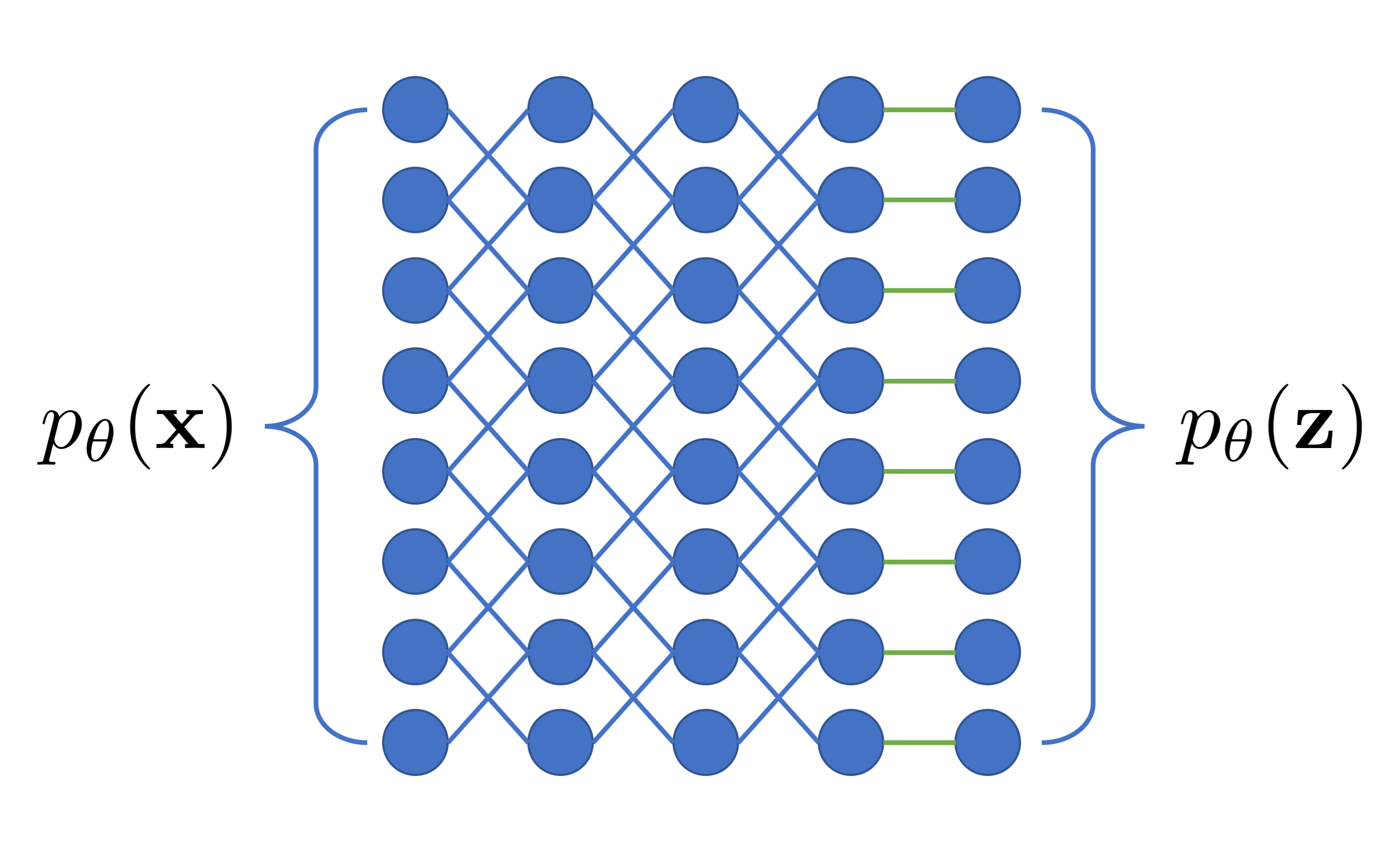}
    \caption{General Flow}
    \label{graphs:flow}
  \end{subfigure}%
  \begin{subfigure}{.33\textwidth}
    \centering
    \includegraphics[width=0.95\linewidth]{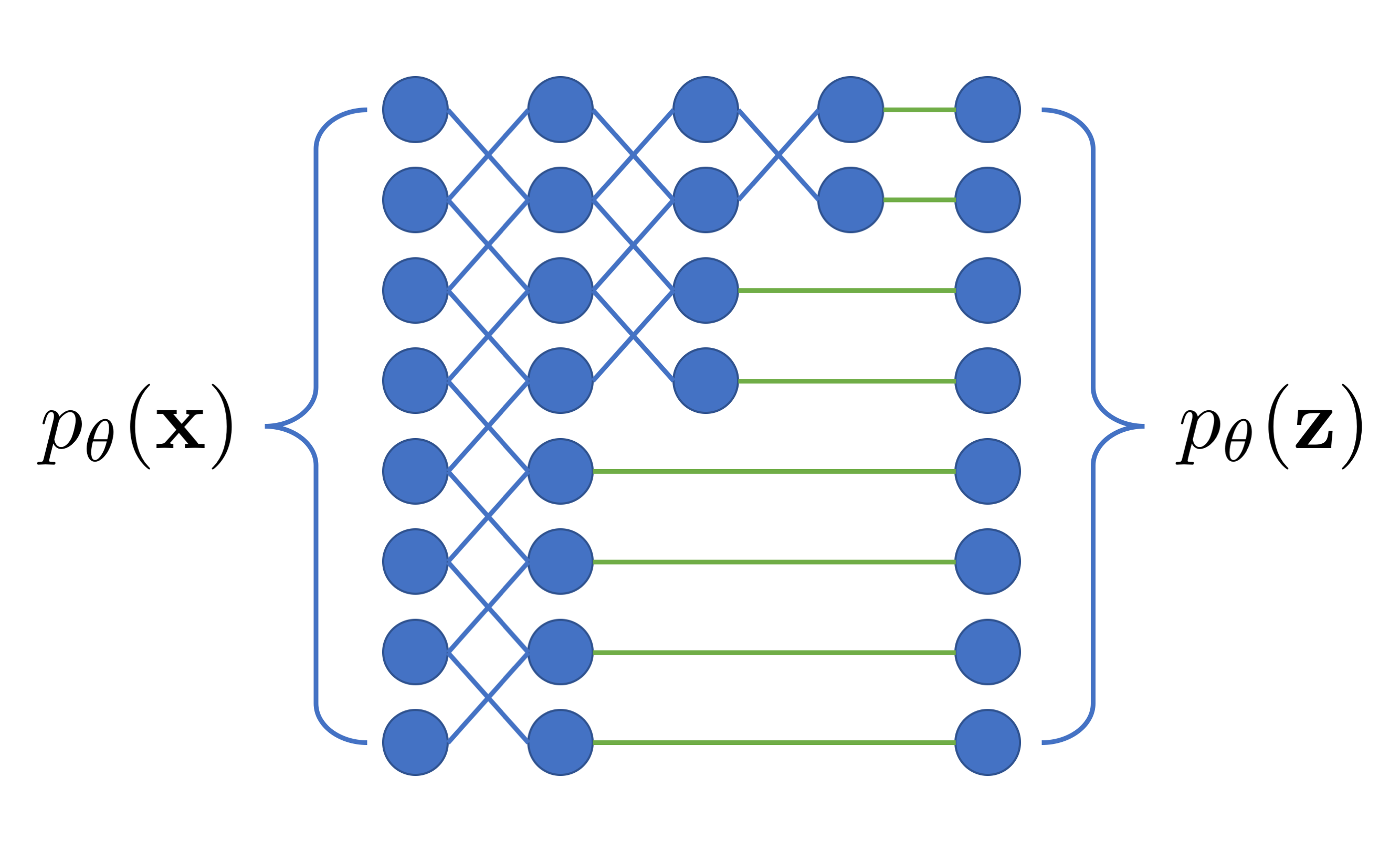}
    \caption{Multi-scale RealNVP}
    \label{graphs:nvp}
  \end{subfigure}
  \begin{subfigure}{.33\textwidth}
    \centering
    \includegraphics[width=0.95\linewidth]{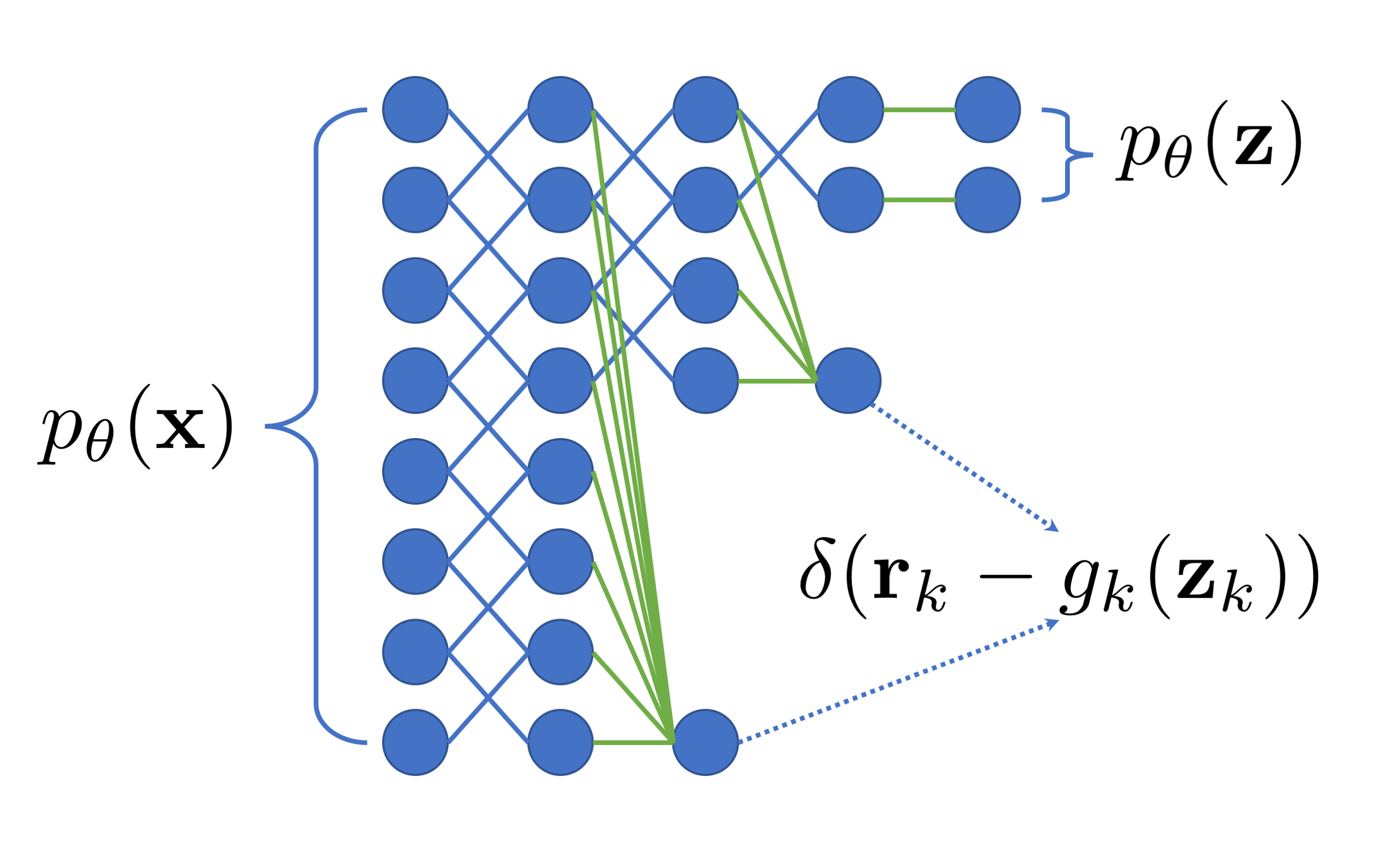}
    \caption{PIE}
    \label{graphs:pie}
  \end{subfigure}
  \caption{Schematic representation of three types of bijective mappings currently used in normalizing 
  flows. The circles represent the variables. The basic invertible mappings are depicted with blue edges. 
  Green edges represent the aggregation of the variables in the objective function. In a general normalizing 
  flow (\subref{graphs:flow}) all the variables are mapped in the same manner and are propagated through 
  the same number of flows. The multi-scale architecture used in RealNVP (\subref{graphs:nvp}) 
  transforms different variables with a different number of flows and afterwards maps them to the same distribution. 
  Our model (\subref{graphs:pie}) progressively discards parts of the variables by hardly constraining their distributions.}
  \label{graphs:fig}
\end{figure}

The method relies on the function $F_{\bm{\theta}}$. 
This choice is challenging by itself. 
The currently known classes of real-value bijectives are limited.
To overcome this issue, we approximate $F_{\bm{\theta}}$ 
with a composition of basic bijectives from certain classes:
\begin{equation}
  F  = F_K \circ F_{k-1} \circ \hdots \circ F_2\circ F_1
\end{equation}
where $F_j = F_j(\cdot | \bm{\theta}_j) \in \mathcal{F}_j,\; j=1\hdots K$. 

Taking into account that a composition of \textit{PIE}s is also a \textit{PIE}, 
we create the final dimensionality reduction mapping from a sequence of \textit{PIE}s:

\begin{equation}
  \mathcal{X}  
  \leftrightarrow \mathcal{Y}_1 
  \leftrightarrow \mathcal{Y}_2 
  \leftrightarrow \dots
  \leftrightarrow \mathcal{Y}_L 
  \leftrightarrow \mathcal{Z}_1 
\end{equation}

such that 
\begin{equation}
D > \dim \mathcal{Y}_1  > \dim \mathcal{Y}_2  > \hdots > \dim \mathcal{Y}_L  > d
\end{equation} 

where $L < D - d$.

Then the log likelihood is represented as 
\begin{equation}
  \label{eq:final_obj}
  \tcboxmath{\log p(\mathbf{x}) \approx \log p(\mathbf{z}) + 
  \sum_{l=1}^{L}\log \mathcal{N}(\mathbf{r}_l | g_l(\mathbf{z}_l), \epsilon_0^2 \mathbf{I}) +
  \sum_{l=1}^{L} \sum_{k=1}^{K_{l}}\log |\det(\mathbf{J}_{kl})| }
\end{equation}
where $\mathbf{J}_{kl}$ is the Jacobian of the $k$-th function of the $l$-th \textit{PIE}.
The approximation error here depends only on $\epsilon$, according to Eq. \ref{eq:delta}.
For the simplicity we will now refer to the whole model as \textit{PIE}. 
The building blocks of this model are \textit{PIE} blocks.

\subsection{Relation to Normalizing Flows} 
If we choose the distribution $p(\mathbf{z})$ in Eq. \ref{eq:final_obj} 
as Standard Gaussian, $g_l(\cdot) = \bm{0},\; \forall l$ and $\epsilon_0 = 1$, 
then the model can be viewed as a Normalizing Flow with a multi-scale architecture \citep{RealNVP} Fig. \ref{graphs:fig}. 
It was demonstrated in \citep{RealNVP} that the model with such an
architecture achieves semantic compression.

\section{Pseudo-Invertible Encoder}
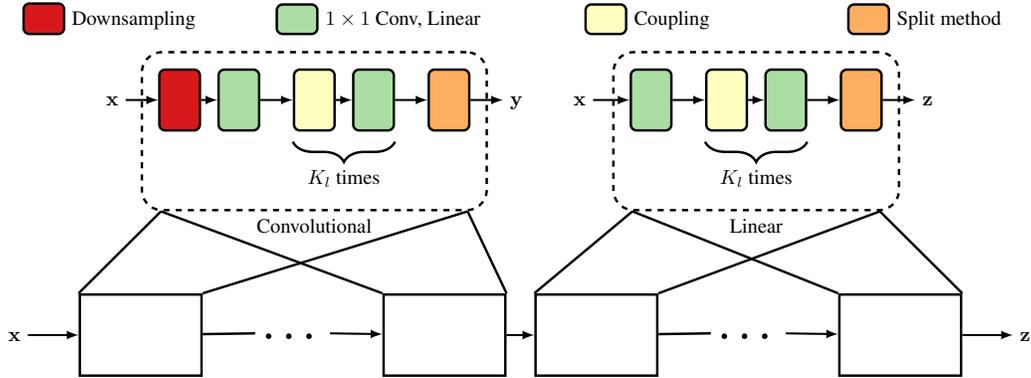
\begin{figure}
    \centering   
    \resizebox{\textwidth}{!}{
        \begin{tikzpicture}[>=latex,text height=1.5ex,text depth=0.25ex]
    \node (conv_input) {$\mathbf{x}$}; 
    \node (conv_pool) [pooling, right=16pt of conv_input] {}; 
    \node (conv_affine) [affine, right=8pt of conv_pool] {}; 
    \node (conv_coupling_x) [coupling, right=16pt of conv_affine] {}; 
    \node (conv_affine_x) [affine, right=8pt of conv_coupling_x] {}; 
    \node (conv_split) [split, right=16pt of conv_affine_x] {}; 
    \node (conv_outpt) [right=16pt of conv_split] {$\mathbf{y}$}; 
    
    \draw [arrow] (conv_input) -- (conv_pool);
    \draw [arrow] (conv_pool) -- (conv_affine);
    \draw [arrow] (conv_affine) -- (conv_coupling_x);
    \draw [arrow] (conv_coupling_x) -- (conv_affine_x);
    \draw [arrow] (conv_affine_x) -- (conv_split);
    \draw [arrow] (conv_split) -- (conv_outpt);
    
    \draw[very thick,decorate,decoration={brace,amplitude=10pt, mirror, raise=5pt}]
        (conv_coupling_x.south west) -- node(conv_curl)[below=15pt]{$K_l$ times} (conv_affine_x.south east);
    
    \node(PIE_conv) [contour, fit=(conv_pool) (conv_split) (conv_curl)] {};
    
    \node (lin_input) [right=18pt of conv_outpt] {$\mathbf{x}$};
    \node (lin_affine) [affine, right=18pt of lin_input] {};
    \node (lin_coupling_x) [coupling, right=16pt of lin_affine] {};
    \node (lin_affine_x) [affine, right=8pt of lin_coupling_x] {};
    \node (lin_split) [split, right=16pt of lin_affine_x] {};
    \node (lin_outpt) [right=16pt of lin_split] {$\mathbf{z}$};
    
    \draw [arrow] (lin_input) -- (lin_affine);
    \draw [arrow] (lin_affine) -- (lin_coupling_x);
    \draw [arrow] (lin_coupling_x) -- (lin_affine_x);
    \draw [arrow] (lin_affine_x) -- (lin_split);
    \draw [arrow] (lin_split) -- (lin_outpt);
    \draw[very thick,decorate,decoration={brace,amplitude=10pt, mirror, raise=5pt}]
        (lin_coupling_x.south west) -- node(lin_curl)[below=15pt]{$K_l$ times} (lin_affine_x.south east);

    \node(PIE_lin) [contour, fit=(lin_affine) (lin_curl) (lin_split)] {};

    \node (conv_1)[block, below=80pt of PIE_conv] at ($(PIE_conv.west)!0.0!(PIE_lin.east)$) {};
    \node (conv_dot)[below=92pt of PIE_conv] at ($(PIE_conv.west)!0.2!(PIE_lin.east)$) {\Huge \dots};
    \node (conv_2)[block, below=80pt of PIE_conv] at ($(PIE_conv.west)!0.4!(PIE_lin.east)$) {};   
    \node (lin_1)[block, below=80pt of PIE_conv] at ($(PIE_conv.west)!0.6!(PIE_lin.east)$) {};
    \node (lin_dot)[below=92pt of PIE_conv] at ($(PIE_conv.west)!0.8!(PIE_lin.east)$) {\Huge \dots};
    \node (lin_2)[block, below=80pt of PIE_conv] at ($(PIE_conv.west)!1.0!(PIE_lin.east)$) {};
    \node (PIE_input)[left=25pt of conv_1] {$\mathbf{x}$};
    \node (PIE_outpt)[right=25pt of lin_2] {$\mathbf{z}$};    

    

    \draw[very thick, -] (PIE_conv.south west) ++(+10pt, 0) -- (conv_1.north west) {}; 
    \draw[very thick, -] (PIE_conv.south east) ++(-10pt, 0) -- (conv_1.north east) {};
    \draw[very thick, -] (PIE_conv.south west) ++(+10pt, 0) -- (conv_2.north west) {}; 
    \draw[very thick, -] (PIE_conv.south east) ++(-10pt, 0) -- (conv_2.north east) {};
    
    \draw[very thick, -] (PIE_lin.south west) ++(+10pt, 0) -- (lin_1.north west) {}; 
    \draw[very thick, -] (PIE_lin.south east) ++(-10pt, 0) -- (lin_1.north east) {};
    \draw[very thick, -] (PIE_lin.south west) ++(+10pt, 0) -- (lin_2.north west) {}; 
    \draw[very thick, -] (PIE_lin.south east) ++(-10pt, 0) -- (lin_2.north east) {};
    
    \draw[arrow] (PIE_input) -- (conv_1);
    \draw[very thick, -] (conv_1) -- (conv_dot);
    \draw[arrow] (conv_dot) -- (conv_2);
    \draw[arrow] (conv_2) -- (lin_1);
    \draw[very thick, -] (lin_1) -- (lin_dot);
    \draw[arrow] (lin_dot) -- (lin_2);
    \draw[arrow] (lin_2) -- (PIE_outpt);
    
    \node[below=0pt of PIE_conv] {Convolutional};
    \node[below=0pt of PIE_lin] {Linear};
    
    
    
    
    \matrix[row sep=4pt, above=of PIE_conv, nodes={anchor=west}] at ($(conv_input)!0.0!(lin_outpt)$){
        \node[pooling, minimum height=15pt] {}; & \node {Downsampling}; \\
    };
    
    \matrix[row sep=4pt, above=of PIE_conv, nodes={anchor=west}] at ($(conv_input)!0.33!(lin_outpt)$){
        \node[affine, minimum height=10pt] {}; & \node {$1 \times 1$ Conv, Linear}; \\
    };
    \matrix[row sep=4pt, above=of PIE_conv, nodes={anchor=west}] at ($(conv_input)!0.66!(lin_outpt)$){
        \node[coupling, minimum height=10pt] {}; & \node {Coupling}; \\       
    };
    \matrix[row sep=4pt, above=of PIE_conv, nodes={anchor=west}] at ($(conv_input)!1.0!(lin_outpt)$){
        \node[split, minimum height=10pt] {}; & \node {Split method}; \\
    };

\end{tikzpicture}
    }
    \caption{Architecture of Pseudo-Invertible Encoder. \textit{PIE} consists of convolutional and linear blocks which can be repeated multiple times, as denoted by the three dots between the block structure at the bottom. Each block has $K_l$ repetitions of the coupling layers and $1 \times 1$ convolutions.}
    \label{fig:PIE}
\end{figure}

This section introduces the basic bijectives for Pseudo-Invertible Encoder (\textit{PIE}).
 We explain what each building bijective consists of and how it fits in the global architecture as 
 shown in Fig. \ref{fig:PIE}.

\subsection{Architecture}
\textit{PIE} is composed of a series of convolutional blocks followed by linear blocks, 
as depicted in Fig. \ref{fig:PIE}. 

The convolutional \textit{PIE} blocks consist of series of coupling layers and $1\times 1$ convolutions. 
We perform invertible downsampling of the image at the beginning of the convolutional block, 
by reducing the spatial resolution and increasing the number of channels, keeping the overall 
number of the variables the same. 
At the end of the convolutional \textit{PIE} block, the split of variables is performed. 
One part of the variables is projected to the residual manifold $\mathcal{R}$ 
while others is feed to the next block.
The linear \textit{PIE} blocks are constructed in the same manner. However, the 
downsampling is not performed and $1\times 1$ convolutions are replaced invertible linear mappings.

\subsection{Coupling layer}
\begin{figure}
	\center
    \begin{subfigure}{1\linewidth}
		\center
		\begin{tikzpicture}[>=latex,text height=1.5ex,text depth=0.25ex]
    \matrix[row sep=0.1cm, column sep=0.5cm] {
        & & \node (x_1) {$\mathbf{x}_1$}; & & &
        \node (mult_1) [operator]{$\boldsymbol{\times}$}; & 
        \node (plus_1) [operator]{$\boldsymbol{+}$}; &
        \node (x_1_anchor) [support]{}; & &
        \node (x_1_anchor2) [support] {}; & & 
        \node (y_1) {$\mathbf{y}_1$}; & & \\

        \node (x) {$\mathbf{x}$}; &
        \node (part) [apply]{$P$}; & & & & & & & & & & & 
        \node (unit) [apply]{$U$}; &
        \node (y) {$\mathbf{y}$}; \\

        & & \node (x_2) {$\mathbf{x}_2$}; &
        \node (x_2_anchor) [support] {}; & &
        \node (x_2_anchor2) [support] {}; & & & & 
        \node (mult_2) [operator]{$\boldsymbol{\times}$}; & 
        \node (plus_2) [operator]{$\boldsymbol{+}$}; &
        \node (y_2) {$\mathbf{y}_2$}; & & \\
    };

    \path[thick, ->]
    (x) edge (part)
    (x_1) edge (mult_1)
    (mult_1) edge (plus_1)
    (plus_1) edge (y_1)
    (x_2) edge (mult_2)
    (mult_2) edge (plus_2)
    (plus_2) edge (y_2)
    (unit) edge (y)
    ;

    \path (x_2_anchor.center) -- node (s_1)[func]{$s_1$} (mult_1.center);
    \path (x_2_anchor2.center) -- node (b_1)[func]{$b_1$} (plus_1.center);
    \path (x_1_anchor.center) -- node (s_2)[func]{$s_2$} (mult_2.center);
    \path (x_1_anchor2.center) -- node (b_2)[func]{$b_2$} (plus_2.center);

    \draw [thick, ->] (part) |- (x_1);
    \draw [thick, ->] (part) |- (x_2);
    \draw [thick, ->] (y_1) -| (unit);
    \draw [thick, ->] (y_2) -| (unit);
    \draw [thick, ->] (x_2_anchor) -- (s_1);
    \draw [thick, ->] (s_1) -- (mult_1);
    \draw [thick, ->] (x_1_anchor) -- (s_2);
    \draw [thick, ->] (s_2) -- (mult_2);
    \draw [thick, ->] (x_2_anchor2) -- (b_1);
    \draw [thick, ->] (x_1_anchor2) -- (b_2); 
    \draw [thick, ->] (b_1) -- (plus_1);
    \draw [thick, ->] (b_2) -- (plus_2);

    \begin{pgfonlayer}{background}
        \node [background, fill=coupling color, fit=(part) (unit) (x_1) (x_2)] {};
    \end{pgfonlayer}

\end{tikzpicture}
		\caption{Forward}    
		\label{fig:coupling_forward}
    \end{subfigure}
    
    \begin{subfigure}{1\linewidth}
    	\center
        \begin{tikzpicture}[>=latex,text height=1.5ex,text depth=0.25ex]
    \matrix[row sep=0.1cm, column sep=0.5cm] {
        & & \node (x_1) {$\mathbf{x}_1$}; &  
        \node (div_1) [operator]{$\boldsymbol{\div}$}; &
        \node (min_1) [operator]{$\boldsymbol{-}$}; & & & &
        \node (x_1_anchor) [support]{}; & &
        \node (x_1_anchor2) [support]{}; &
        \node (y_1) {$\mathbf{y}_1$}; & & \\

        \node (x) {$\mathbf{x}$}; &
        \node (unit) [apply] {$P^{-1}$}; & & & & & & & & & & & 
        \node (part) [apply] {$U^{-1}$}; & 
        \node (y) {$\mathbf{y}$}; \\

        & & \node (x_2) {$\mathbf{x}_2$}; & & 
        \node (x_2_anchor) [support]{}; & & 
        \node (x_2_anchor2) [support]{}; & 
        \node (div_2) [operator]{$\boldsymbol{\div}$}; &
        \node (min_2) [operator]{$\boldsymbol{-}$}; & & &
        \node (y_2) {$\mathbf{y}_2$}; & & \\
     };

    \path (x_2_anchor2.center) -- node (b_1)[func]{$b_1$} (min_1.center);
    \path (x_2_anchor.center) -- node (s_1)[func]{$s_1$} (div_1.center);
    \path (x_1_anchor2.center) -- node (b_2)[func]{$b_2$} (min_2.center);
    \path (x_1_anchor.center) -- node (s_2)[func]{$s_2$} (div_2.center);

    \draw [thick, ->]
    (y) edge (part)
    (y_1) edge (min_1)
    (min_1) edge (div_1)
    (div_1) edge (x_1)
    (y_2) edge (min_2)
    (min_2) edge (div_2)
    (div_2) edge (x_2)
    (unit) edge (x)
    ;

    \draw [thick, ->] (part) |- (y_1);
    \draw [thick, ->] (part) |- (y_2);
    \draw [thick, ->] (x_1) -| (unit);
    \draw [thick, ->] (x_2) -| (unit);
    \draw [thick, ->] (x_1_anchor) -- (s_2);
    \draw [thick, ->] (s_2) -- (div_2);
    \draw [thick, ->] (x_2_anchor) -- (s_1);
    \draw [thick, ->] (s_1) -- (div_1);
    \draw [thick, ->] (x_1_anchor2) -- (b_2);
    \draw [thick, ->] (b_2) -- (min_2);
    \draw [thick, ->] (x_2_anchor2) -- (b_1);
    \draw [thick, ->] (b_1) -- (min_1);

    \begin{pgfonlayer}{background}
        \node [background, fill=coupling color, fit=(part) (unit) (x_1) (x_2)] {};
    \end{pgfonlayer}
\end{tikzpicture}
        \caption{Inverse}
        \label{fig:coupling_inverse}
    \end{subfigure}
    
    \caption{
    Structure of the coupling block. $P$ partitions the input into two groups of equal length. 
    $U$ unites these groups together. In the inverse $P^{-1}$ and $U^{-1}$ are 
    the inverse of these operations respectively.}
    \label{fig:coupling}
\end{figure}
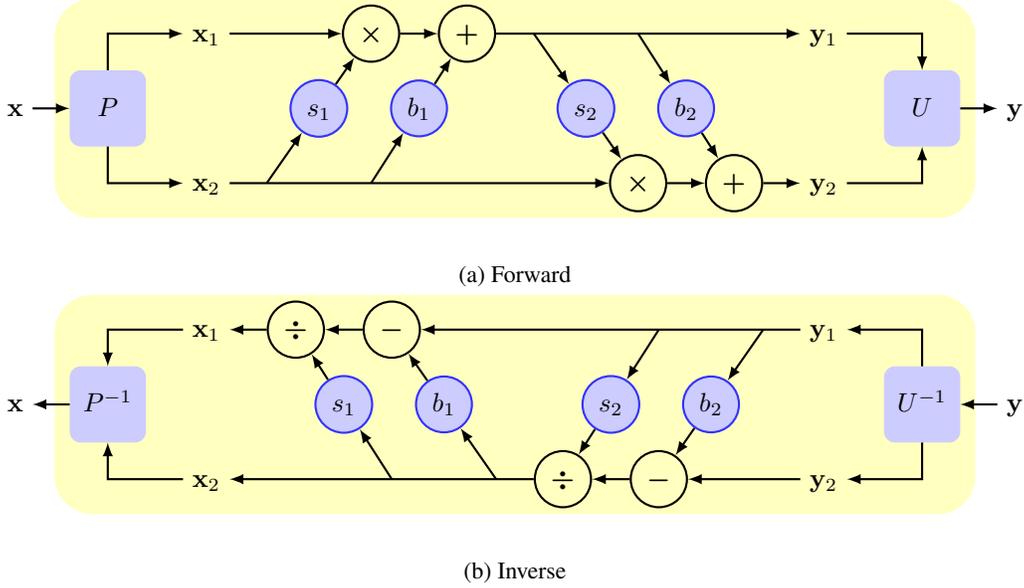

In order to enhance the flexibility of the model, we utilize affine coupling layers Fig. \ref{fig:coupling}.
We modify the version, introduced in \citep{RealNVP}.

Given input data $\mathbf{x}$, the output $\mathbf{y}$ is 
obtained by using the mapping:

\begin{equation}
    \label{eq:affine}
    \begin{cases}
        \mathbf{y}_1 = s_1(\mathbf{x}_2) \odot \mathbf{x}_1 + b_1(\mathbf{x}_2) \\
        \mathbf{y}_2 = s_2(\mathbf{y}_1) \odot \mathbf{x}_2 + b_2(\mathbf{y}_1)
    \end{cases}
    \Longleftrightarrow
    \begin{cases}
        \mathbf{x}_2 = (\mathbf{y}_2 - b_2(\mathbf{y}_1)) / s_2(\mathbf{y}_1)\\
        \mathbf{x}_1 = (\mathbf{y}_1 - b_1(\mathbf{x}_2)) / s_1(\mathbf{x}_2)
    \end{cases}
\end{equation}

Here multiplication $\odot$ and division are performed element-wisely. 
The scalings $s_1, s_2$ and the biases $b_1, b_2$ are functions, 
parametrized by neural networks. Invertibility is not required for this functions.
$\mathbf{x}_1, \mathbf{x}_2$ are non-intersecting 
partitions of $\mathbf{x}$. 
For convolutional blocks, we partition the tensors by splitting them into halves along the channels.
In the case of the linear blocks, we just split the features into halves.

The log determinant of the Jacobian of the coupling layer is given by:

\begin{equation*}
    \label{eq:logdet_affine}
    \log \Big| \det \Big(  \frac{\partial F_{\bm{\theta}}}{\partial \mathbf{x}^T} \Big)\Big| = 
    \text{sum}(\log |s_1| ) +  \text{sum}(\log |s_2| )
\end{equation*}

where $\log|\cdot|$ is calculated element-wisely.

\subsection{Invertible $1 \times 1$ Convolution and Linear Transformation}
The affine couplings operate on non-intersecting parts of the tensor. 
In order to capture various correlations between channels and features, a
different mechanism of channel permutations was proposed. 
\citep{Glow} demonstrated that invertible $1 \times 1$ convolutions perform better than fixed permutations and reversing of the order of channels \citep{RealNVP}.

We parametrize Invertible $1 \times 1$ Convolutions and invertible linear mappings with Householder Matrices \citep{Householder}.
Given the vector $\mathbf{v}$, the Householder Matrix is computed as:
\begin{equation}
	\mathbf{H}(\mathbf{v}) = \mathbf{I} - 2 \frac{\mathbf{v}\mathbf{v}^T}{\mathbf{v}^T\mathbf{v}}
    \label{eq:hh_transformation}
\end{equation}

The obtained matrix is orthogonal. Therefore, its inverse is just its transpose, 
which makes the computation of the inverse easier compared to \citep{Glow}.
The log determinant of the Jacobian of such transformation is equal to $0$.

\subsection{Downsampling}
We use invertible downsampling to progressively reduce the spatial 
size of the tensor and increase the number of its channels. 
The downsampling with the checkerboard patterns \citep{iRevNet,RealNVP} transforms the tensor of size
$C \times H \times W$ into a tensor of size $4C \times \frac{H}{2} \times \frac{W}{2}$, 
where $H, W$ are the height and the width of the image, and $C$ is the number of the channels.
The log determinant of the Jacobian of Downsampling is $0$ as it just performs permutation.
\subsection{Split}
\begin{figure}
    \resizebox{\linewidth}{!}{
    \begin{subfigure}{0.5\linewidth}
    	\center
		\begin{tikzpicture}[text height=1.5ex,text depth=0.25ex]
    \matrix[row sep=0.1cm, column sep=0.5cm] {

        & & \node (x_tilde) {$\mathbf{z}$}; & & & \\
        \node (x) {$\mathbf{x}$}; &
        \node (part) [apply]{$P$}; & & &
        \node (empty){};  &
        \node (y) {$\mathbf{z}$}; \\
        & & \node (r) {$\mathbf{r}$}; & \node (r_anchor) {$\delta (\mathbf{r}-g(\mathbf{z}))$};& & \\
    };

    \path[thick, ->]
    (x) edge (part);

    \draw [thick, ->] (part) |- (x_tilde);
    \draw [thick, ->] (part) |- (r);
    \draw [thick, ->] (x_tilde) -| (empty.center) -- (y);
    \draw [thick, ->] (r) -- (r_anchor);

    \begin{pgfonlayer}{background}
        \node [background, fill=split color, fit=(part) (x_tilde) (r) (empty)] {};
    \end{pgfonlayer}
\end{tikzpicture}
		\caption{Forward}
		\label{fig:split_forward}
    \end{subfigure}
    \begin{subfigure}{0.5\linewidth}
    	\center
		\begin{tikzpicture}[text height=1.5ex,text depth=0.25ex]
    \matrix[row sep=0.1cm, column sep=0.5cm,] {
        & & \node (x_tilde) {$\mathbf{z}$}; & & & \\
        \node (x) {$\mathbf{x}$}; &
        \node (part) [apply]{$P^{-1}$}; & & &
        \node (empty){};  &
        \node (y) {$\mathbf{z}$}; \\
        & & \node (r) {$\mathbf{r}=g(\mathbf{z})$}; & \node (r_anchor) {$\;\;\;\,\,$};& & \\
    };

    \path[thick, <-]
    (x) edge (part);

    \draw [thick, <-] (part) |- (x_tilde);
    \draw [thick, <-] (part) |- (r);
    \draw [thick, <-] (x_tilde) -| (empty.center) -- (y);
    \draw [thick, <-] (r) -- (r_anchor);
    \draw [thick, <-] (r) -| (empty.center);

    \begin{pgfonlayer}{background}
        \node [background, fill=split color, fit=(part) (x_tilde) (r) (empty)] {};
    \end{pgfonlayer}
\end{tikzpicture}
		\caption{Inverse} 
		\label{fig:split_inverse}
    \end{subfigure}
    }
    \caption{
        Structure of the split method. $P$ partitions the input into two subsamples. $P^{-1}$ unites these subsamples together.}
    \label{fig:split}
\end{figure}
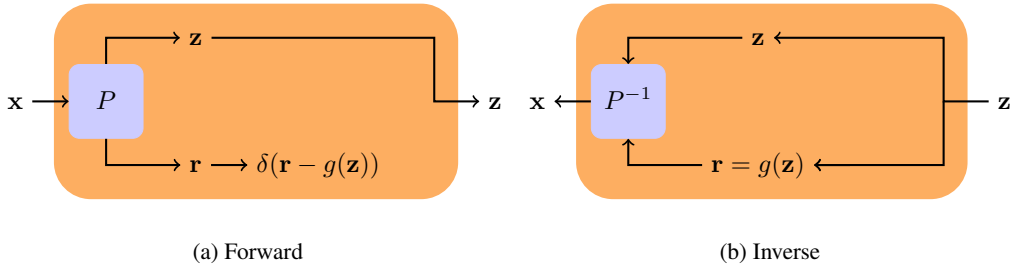

All the discussed blocks transform the data while preserving its dimensionality. 
Here we introduce Split block Fig. \ref{fig:split}, which is responsible for the \textit{projection}, 
\textit{restrictions} and \textit{extension}, described in Section 3. 
It reduces dimensionality of the data by splitting the variables 
into two non-intersecting parts $\mathbf{z}, \mathbf{r}$ of dimensionalities $d$ and $D-d$, respectively.
$\mathbf{z}$ is kept and is processed by subsequent blocks. $\mathbf{r}$ is constrained to match 
$\mathcal{N}(\mathbf{r}| g(\mathbf{z}), \epsilon^2_0\mathbf{I})$. The mappings is defined as 

\begin{equation}
	\label{eq:split}
	\begin{cases}
		\mathbf{z} = \left. \mathbf{x} \right|_{\mathbb{R}^d} \\
		\mathbf{r} \rightarrow \mathcal{N}(\mathbf{r}| g(\mathbf{z}), \epsilon^2_0\mathbf{I})
	\end{cases}
	\Longleftrightarrow
	\mathbf{x} = [\mathbf{z}, g(\mathbf{z})]
\end{equation}

\section{Experiments}

\begin{figure}
    \centering
    \begin{subfigure}[t]{\linewidth}
        \begin{minipage}[c]{0.01\textwidth}
        \caption{ }
        \label{fig:reconstruction}
        \end{minipage}\hfill
        \begin{minipage}[c]{0.99\textwidth}
        \centering
        \includegraphics[width=.3\textwidth]{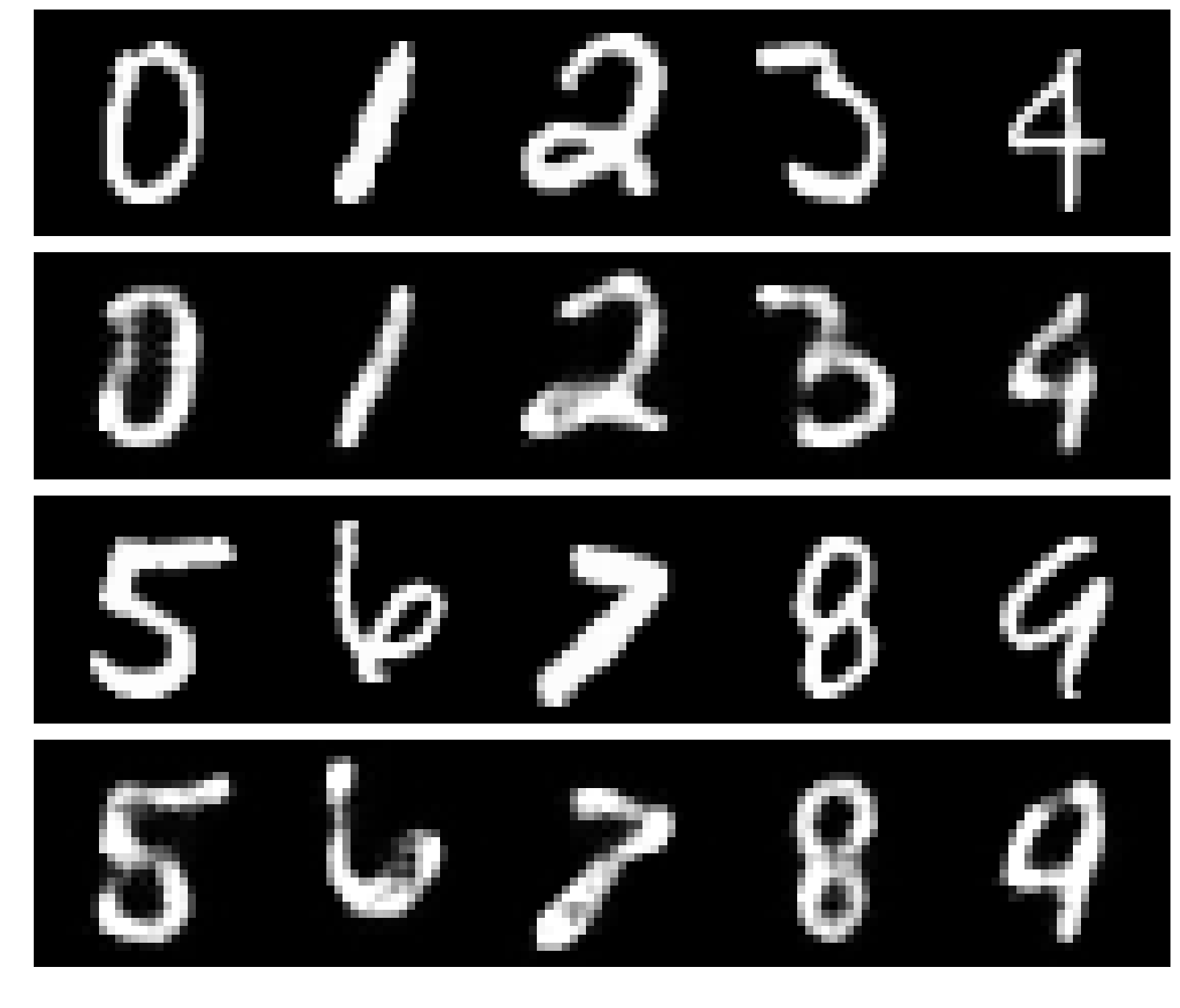}
        \includegraphics[width=.3\textwidth]{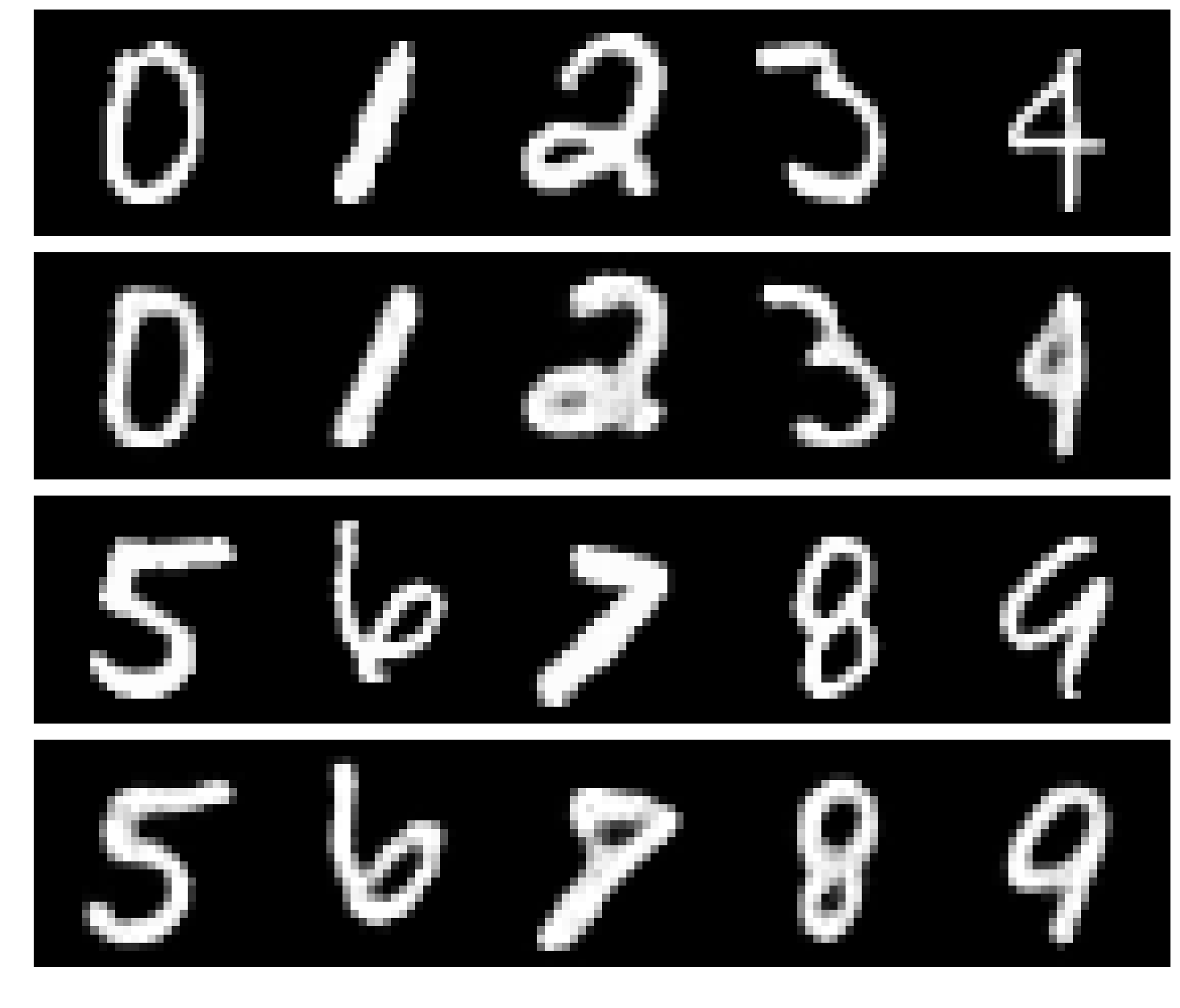}
        \includegraphics[width=.3\textwidth]{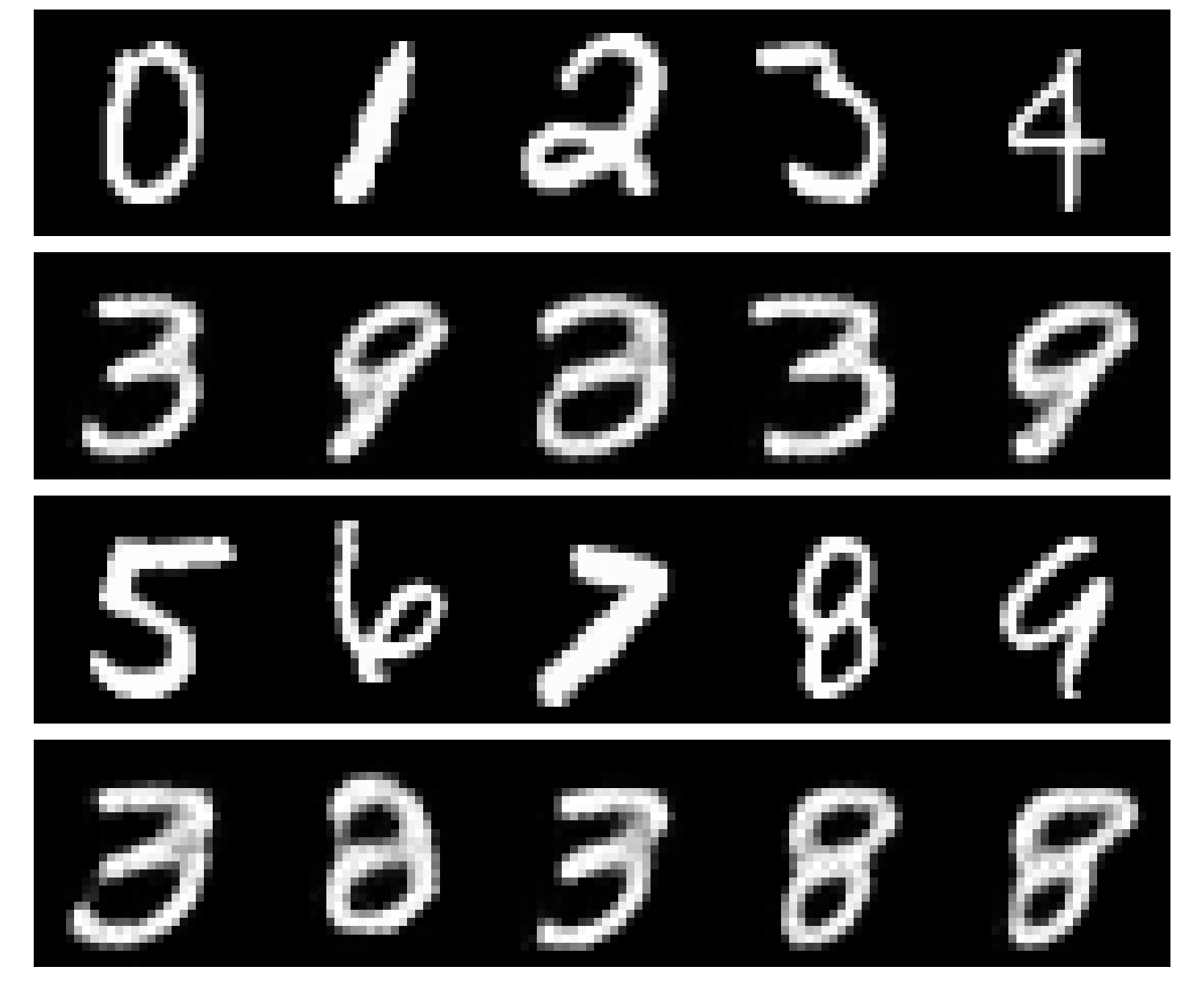}
        \end{minipage}
    \end{subfigure}
    \begin{subfigure}[t]{\linewidth}
        \begin{minipage}[c]{0.01\textwidth}
        \caption{ }
        \label{fig:sampling_N(0,1)}
        \end{minipage}\hfill
        \begin{minipage}[c]{0.99\textwidth}
        \centering
        \includegraphics[width=.3\textwidth]{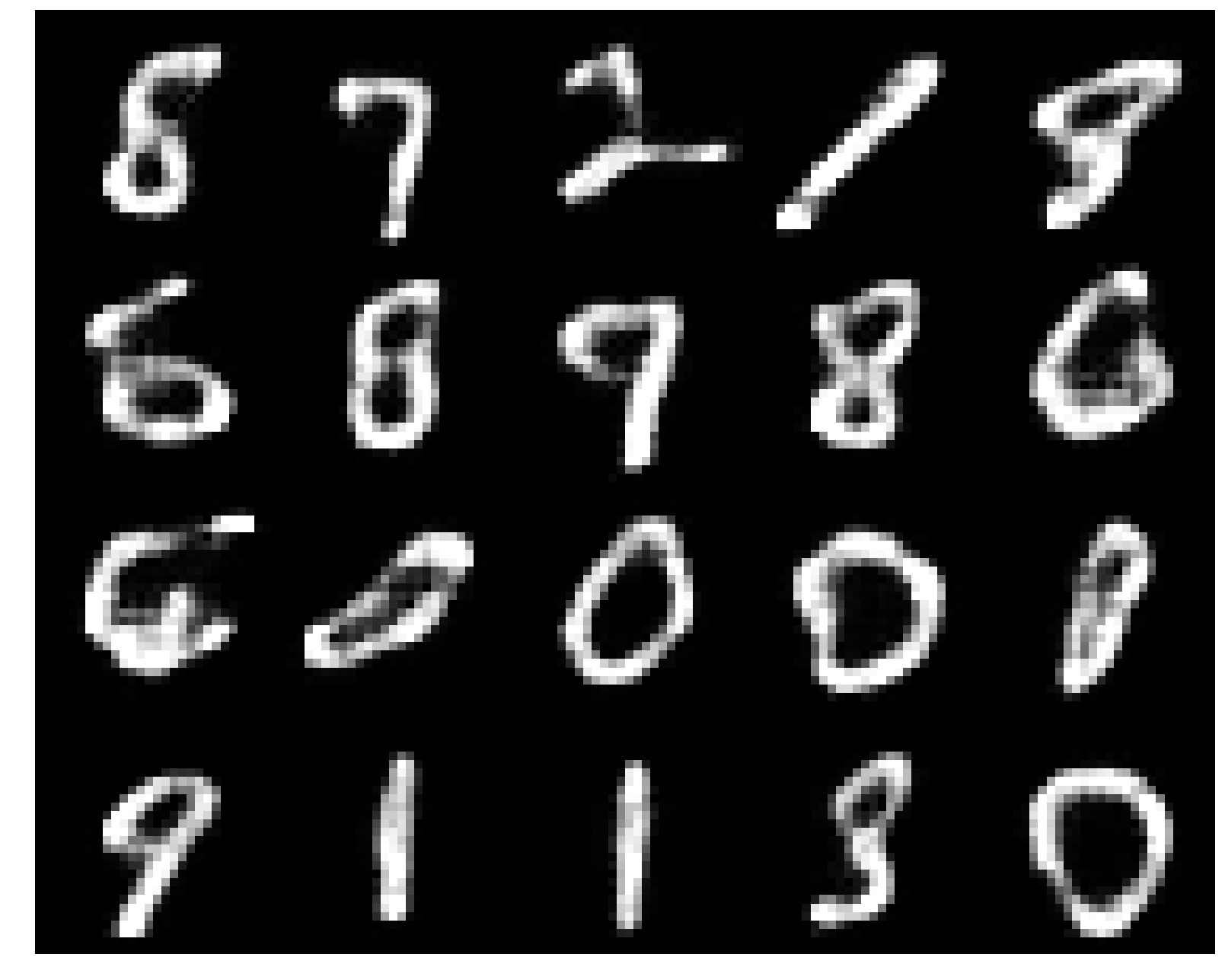}
        \includegraphics[width=.3\textwidth]{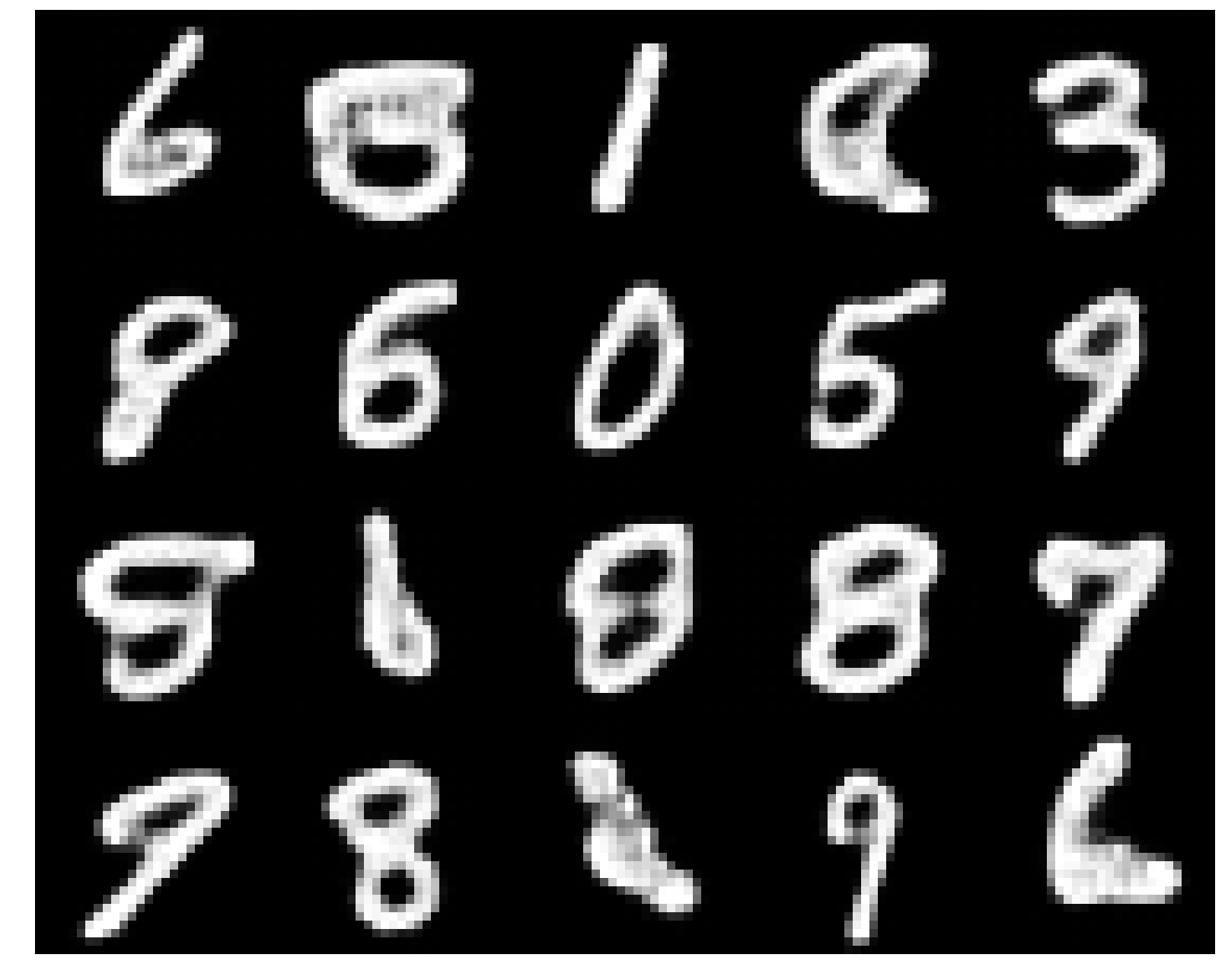}
        \includegraphics[width=.3\textwidth]{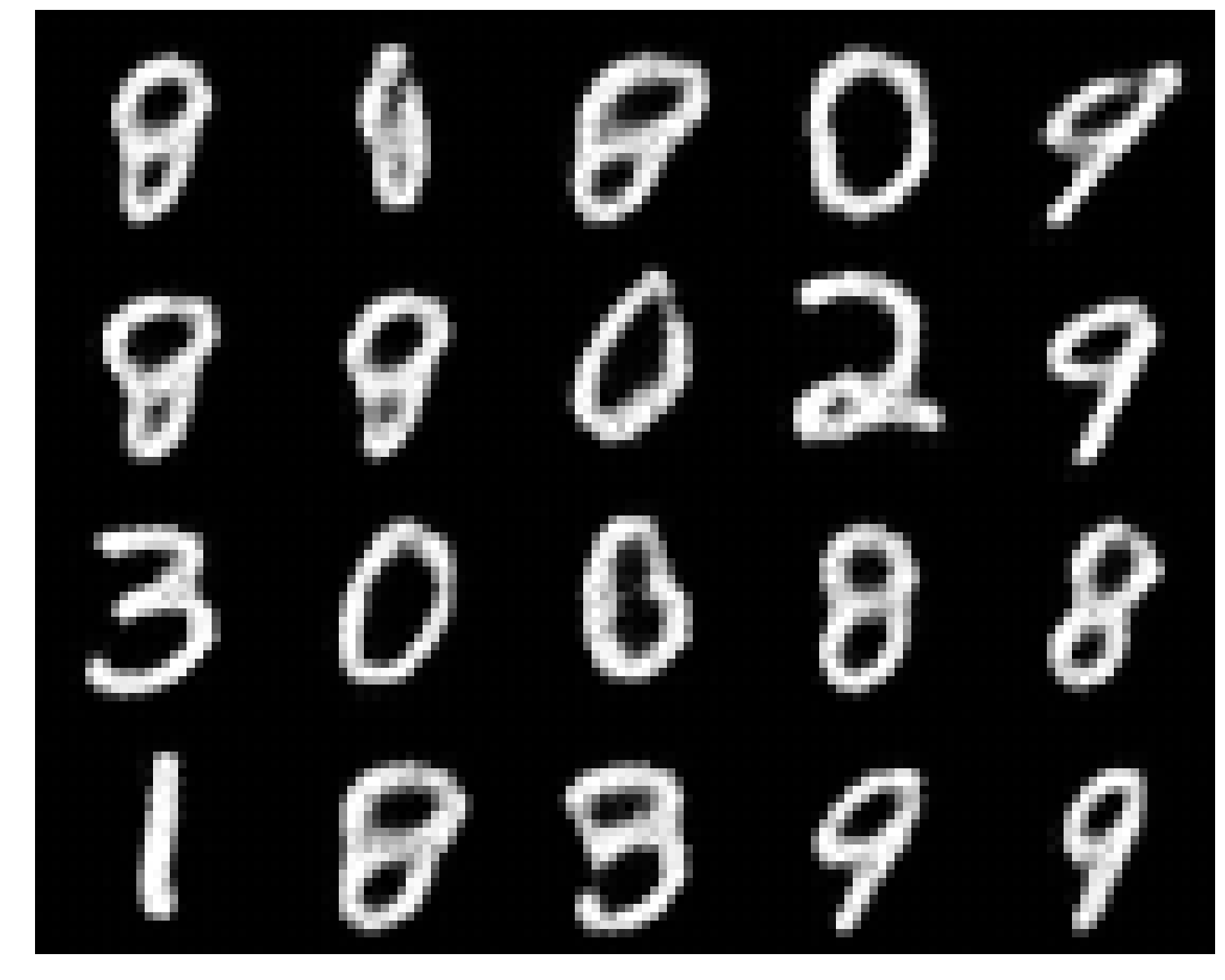}
        \end{minipage}
    \end{subfigure}
    \begin{subfigure}[t]{\linewidth}
        \begin{minipage}[c]{0.01\textwidth}
        \caption{ }
        \label{fig:sampling_N(0,0.1)}
        \end{minipage}\hfill
        \begin{minipage}[c]{0.99\textwidth}
        \centering
        \includegraphics[width=.3\textwidth]{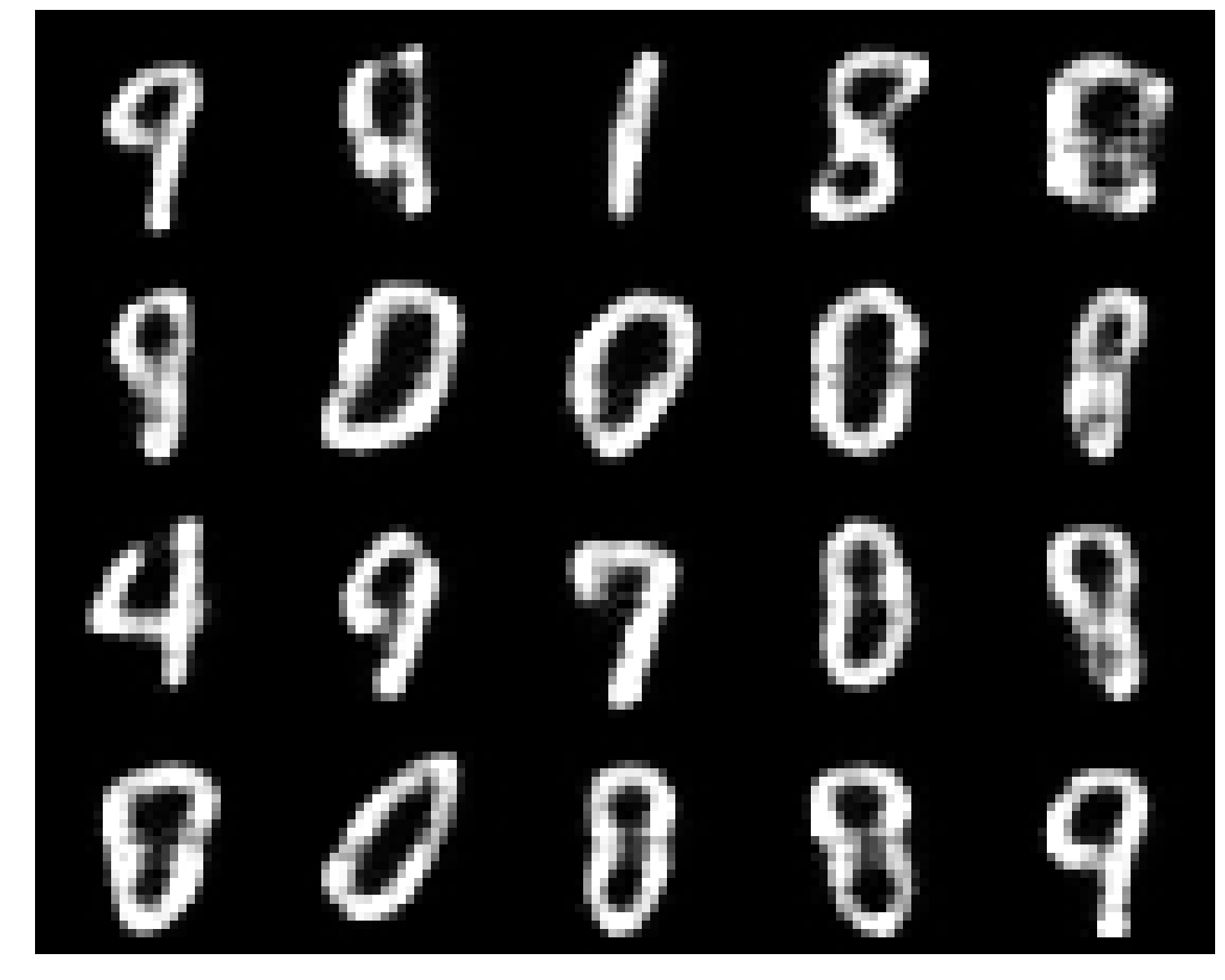}
        \includegraphics[width=.3\textwidth]{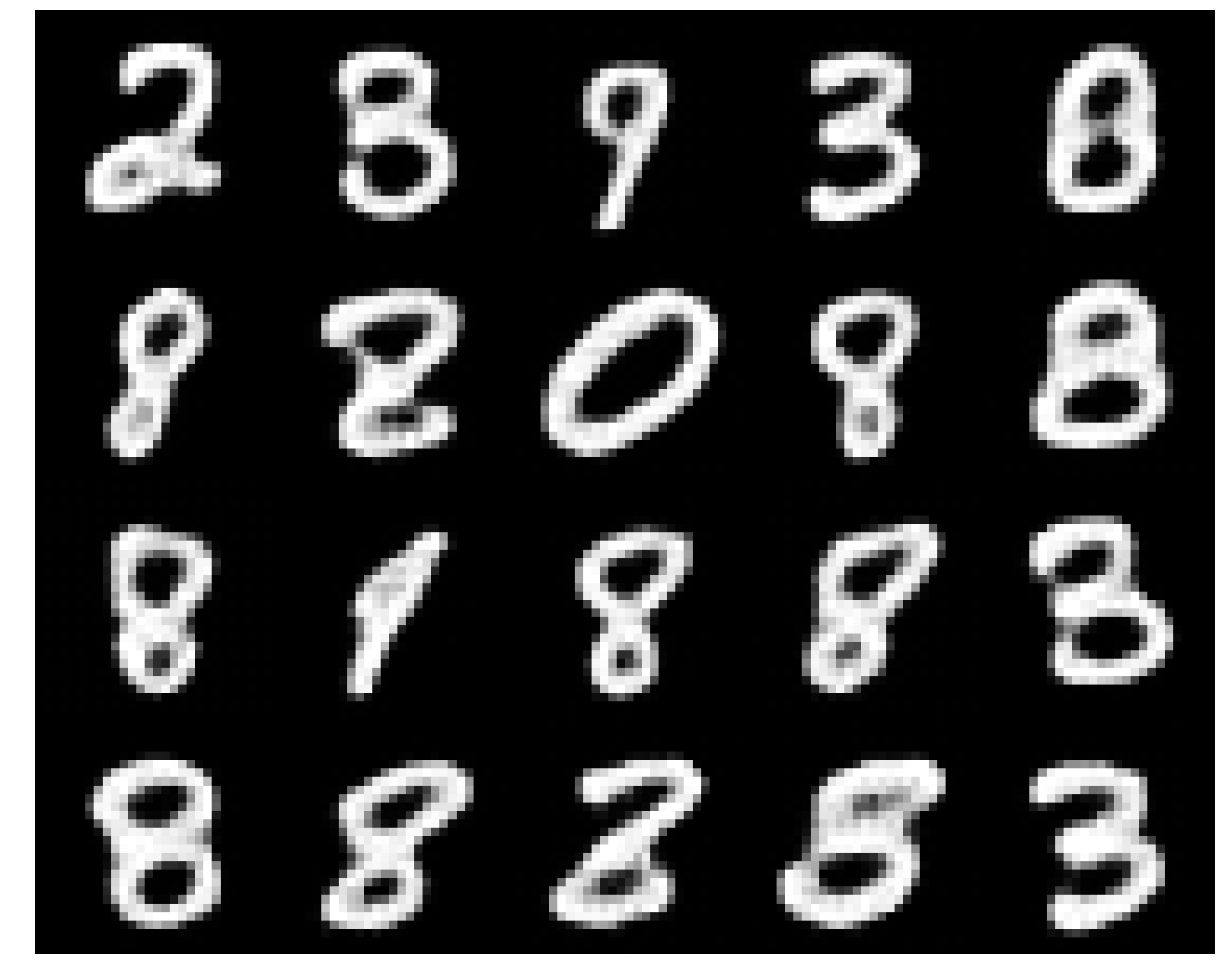}
        \includegraphics[width=.3\textwidth]{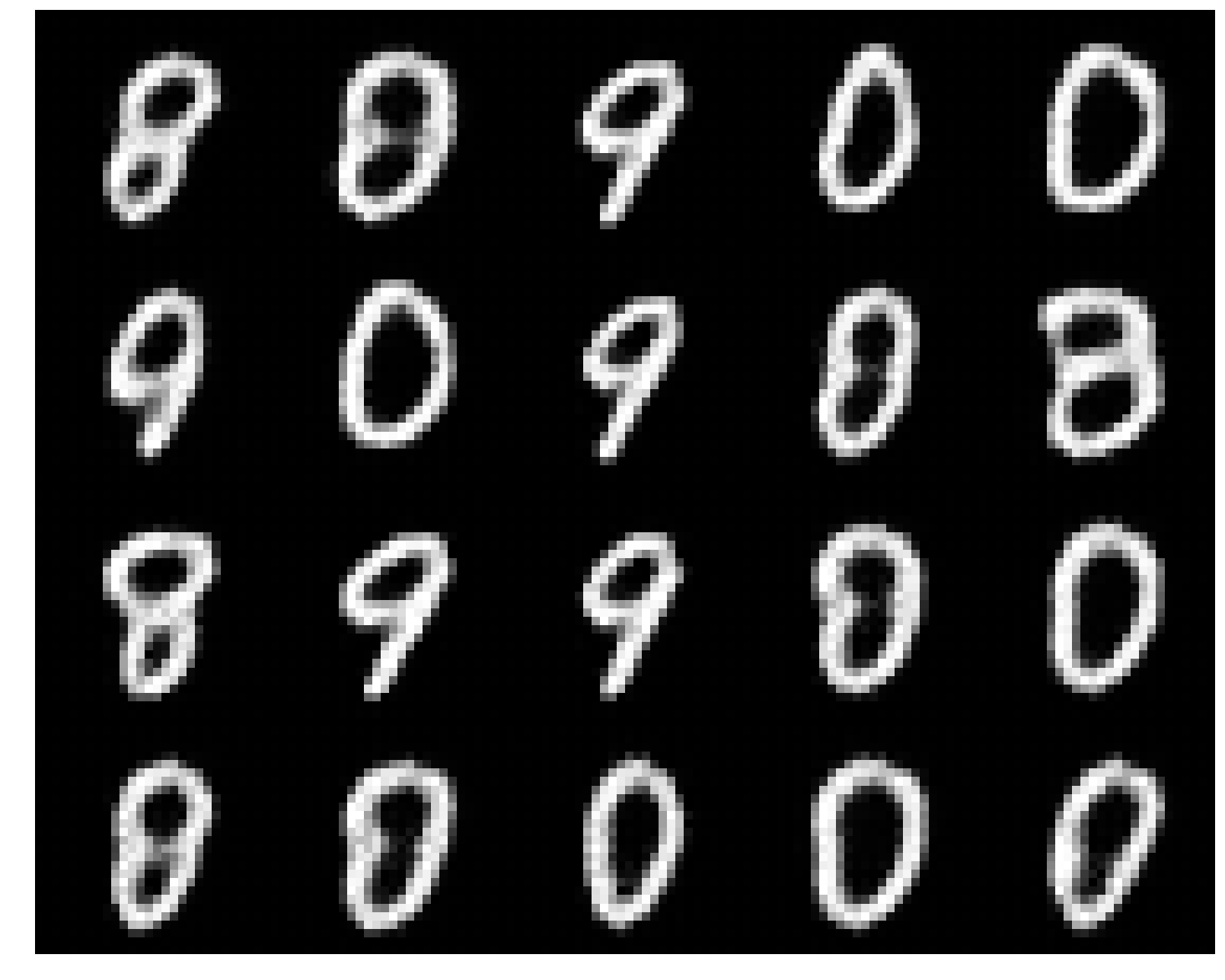}
        \end{minipage}
    \end{subfigure}
    \begin{subfigure}[t]{\linewidth}
        \begin{minipage}[c]{0.01\textwidth}
        \caption{ }
        \label{fig:linear_interpolation}
        \end{minipage}\hfill
        \begin{minipage}[c]{0.99\textwidth}
        \centering
        \includegraphics[width=.3\textwidth]{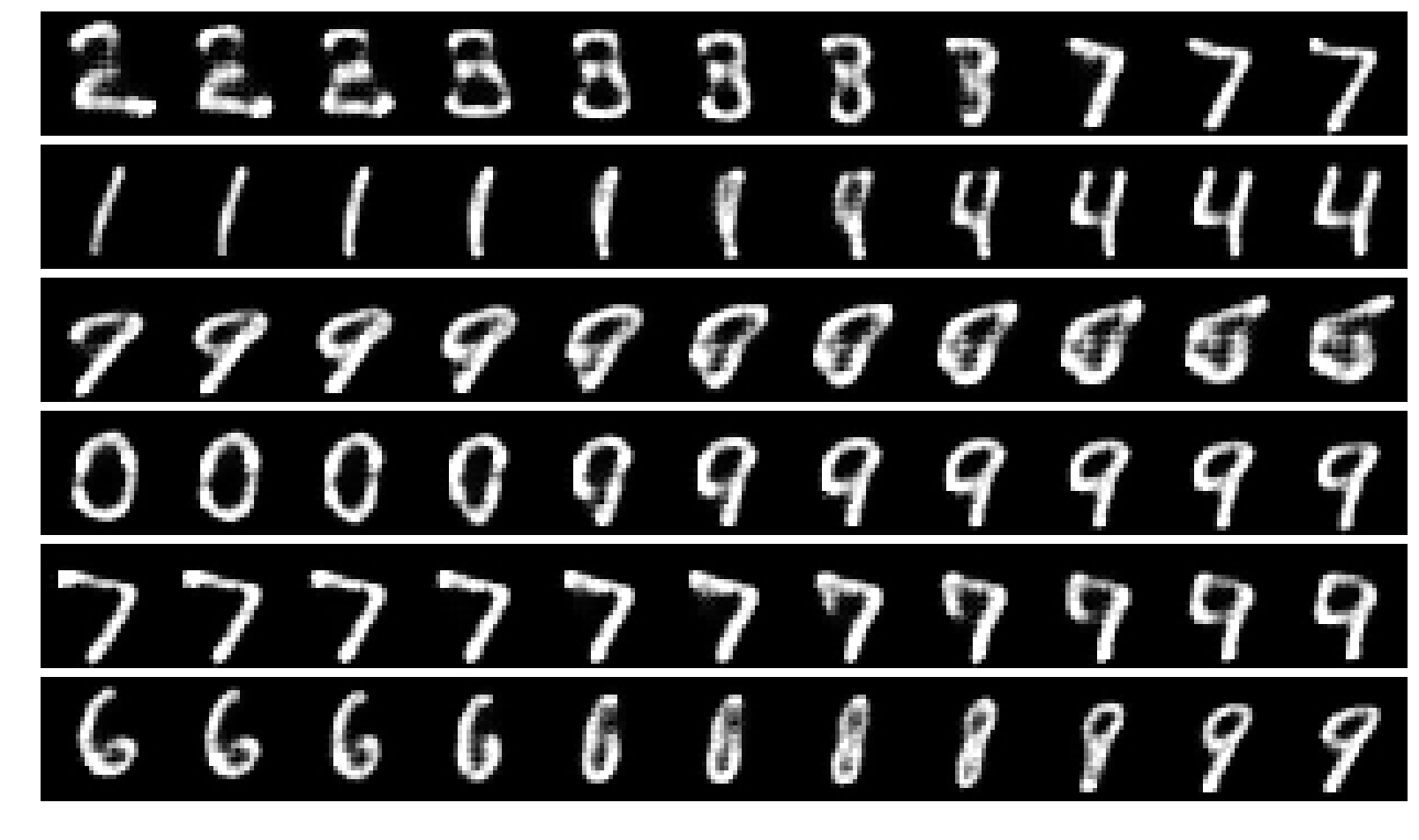}
        \includegraphics[width=.3\textwidth]{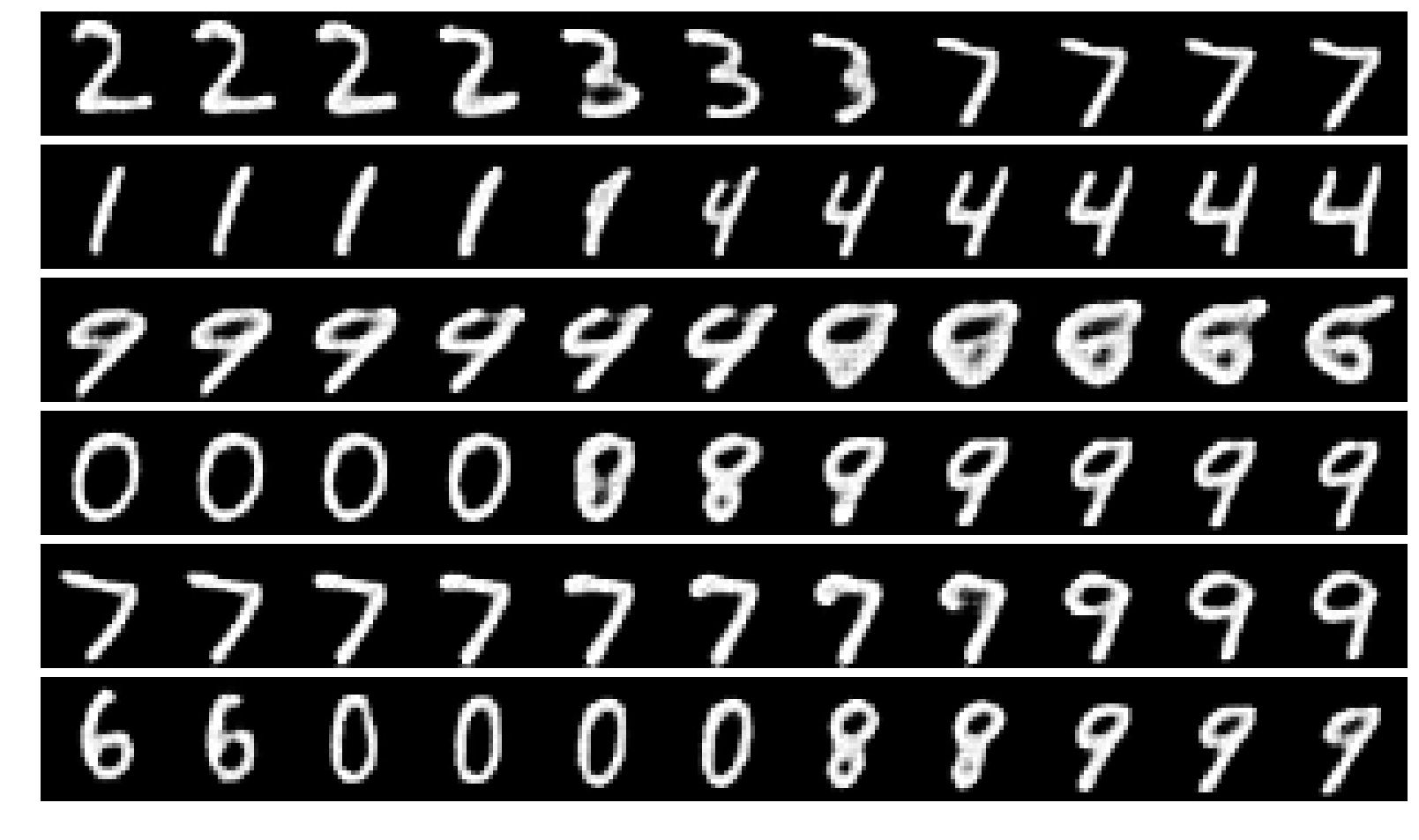}
        \includegraphics[width=.3\textwidth]{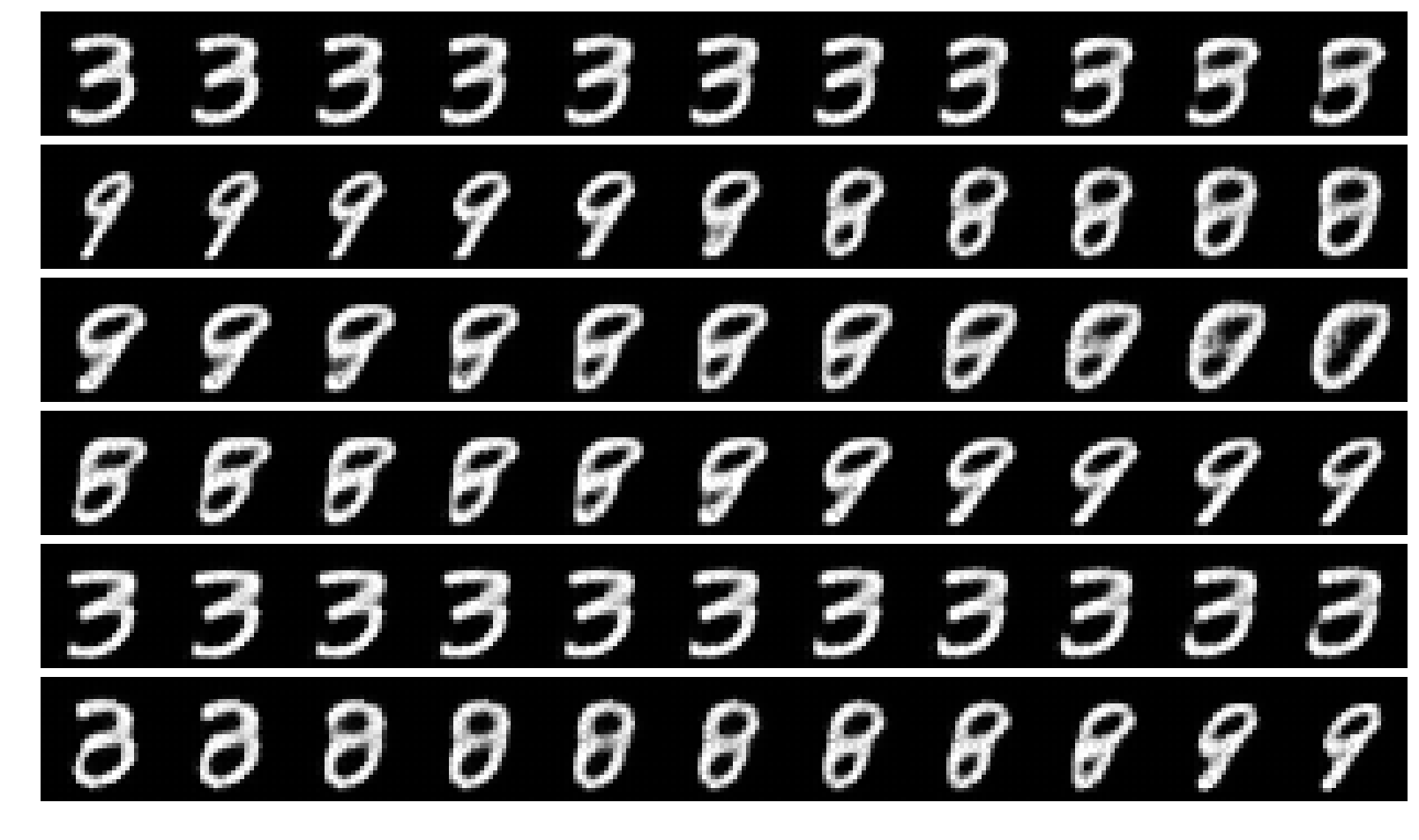}
        \end{minipage}
    \end{subfigure}
    \begin{subfigure}[t]{\linewidth}
        \begin{minipage}[c]{0.01\textwidth}
        \caption{ }
        \label{fig:UMAP}
        \end{minipage}\hfill
        \begin{minipage}[c]{0.99\textwidth}
        \centering
        \includegraphics[width=.3\textwidth]{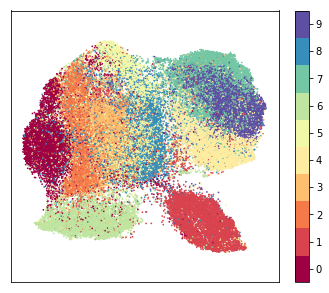}
        \includegraphics[width=.3\textwidth]{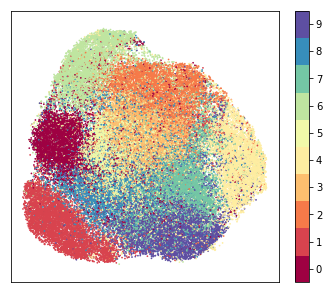}
        \includegraphics[width=.3\textwidth]{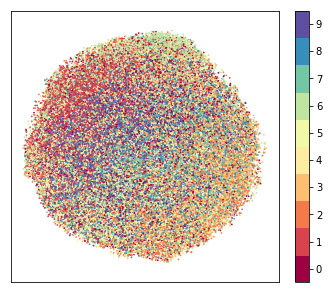}
        \end{minipage}
    \end{subfigure}
    \begin{tabular}{p{4cm} p{3.5cm} p{2cm}}
        $\epsilon ^2 = 0.01$ & $\epsilon ^2 = 0.1$ & $\epsilon ^2 = 1.0$ \\ 
    \end{tabular}

    \caption{Experiments on the MNIST dataset with $\dim (\mathbf{z}) = 10$. (a) shows reconstructions on the test data. Row 1 and 3 are the original images, row 2 and 4 are reconstructions from $z$-space. (b) and (c) both show samples from $z$-space. (b) is sampled from $\mathcal{N}(0,1)$, (c) is sampled from $\mathcal{N}(0,0.5)$. (d) is a linear interpolation between a picture on the left and the right of the image. All digits shown are reconstructed from $z$-space. At last (e) shows UMAP from $\dim (\mathbf{z}) = 10$.}
    \label{fig:mnist_experiment}
\end{figure}

\subsection{Manifold learning}
\label{sec:manifold_learning}
For this experiment, we trained a Gaussian \textit{PIE} on the MNIST digits dataset. We build a \textit{PIE} with 2 convolutional blocks, each splitting the data in the last layer to 50\% of the input size. Next, we add three linear blocks to the \textit{PIE}, reducing the dimensions to 64, 10 and the last block does not reduce the dimensions any further. For each affine transformation, we use the three biggest possible Householder reflections. For this experiment, we set $K_l$ equal to 3. Optimization is done with the Adam optimizer \citep{Kingma2014}. The model diminishes the number of dimensions from $784$ to $10$.

This experiment shows the ability of \textit{PIE} to learn a manifold with three different constraints; $\epsilon ^2 = 0.01$, $\epsilon ^2 = 0.1$ and $\epsilon ^2 = 1.0$. The results are shown in Fig. \ref{fig:mnist_experiment}. As the constraint gets too loose, as shown in the right column, the model is not able to reconstruct anymore (Fig. \ref{fig:reconstruction}). Lower values for $\epsilon ^2$ perform better in terms of reconstruction. Too low values, however, sample fuzzy images (Fig. \ref{fig:sampling_N(0,1)}). Narrowing down the distribution to sample from increases the models probability to produce accurate images. This is shown in Fig. \ref{fig:sampling_N(0,0.1)} where samples are taken from $\mathcal{N}(0, 0.5)$. For both $\epsilon ^2 = 0.01$ and $\epsilon ^2 = 0.1$ reconstructed images are more accurate. 

Fig. \ref{fig:linear_interpolation} shows for each model a linear interpolation from one latent space to another. Both lower values of $\epsilon ^2$ ($0.01$, $0.1$) show digits that are quite accurate. When the constraint is loosened to $\epsilon = 1.0$ the interpolation is unable to show distinct values. 

This experiment shows that tightening the constraint by decreasing $\epsilon ^2$ increases the power of the manifold learned by the model. This is shown again in Fig. \ref{fig:UMAP} where we diminished the number of dimensions even further from $\mathbb{R}^{10}$ to $\mathbb{R}^2$ utilizing UMAP \citep{McInnes2018}. With $\epsilon ^2 = 1.0$ UMAP created a manifold with good Gaussian distribution. However, from the manifold created by the \textit{PIE} it was not able to separate distinct digits from each other. Tightening the constraint with a lower $\epsilon ^2$ moves the manifold created by UMAP further away from a Gaussian distribution, while it is better able to separate classes from each other.

\subsection{Image sharpness}
It is a well-known problem in VAEs that generated images are blurred. WAE \citep{Tolstikhin2018} improves over VAEs by utilizing Wasserstein distance function. To test the sharpness of generated images we convolve the grey-scaled images with the Laplace filter. This filter acts as an edge detector. We compute the variance of the activations and average them over 10000 sampled images. If an image is blurry, it means there are fewer edges and thus more activations will be close to zero, leading to a smaller variance. In this experiment, we compare the sharpness of the images generated by \textit{PIE} with WAE, VAE, and the sharpness of the original images. For VAE and WAE, we take the architecture as described in \citep{Radford2015}. For \textit{PIE} we take the architecture as described in section \ref{sec:manifold_learning}.

Table \ref{tab:smootheness} shows the results for this experiment. \textit{PIE} outperforms both VAE and WAE in terms of the sharpness of generated images. Images generated by \textit{PIE} are even sharper than the original images from the MNIST dataset. An explanation for this is the use of a checkerboard pattern in the downsampling layer of the \textit{PIE} convolutional block. With this technique, we capture intrinsic properties of the data and are thus able to reconstruct sharper images.

\begin{table}[]
    \centering
    \begin{tabular}{c|c}
         & Sharpness \\ \hline
        True & $0.18$ \\
        VAE & $0.08$ \\
        WAE & $0.07$ \\
        \textit{PIE} & $\mathbf{0.49}$ \\
    \end{tabular}
    \caption{Results for experiment on sharpness on three different models and original images. For all three models a sample of 8 dimensions was taken. The generated images where convolved with Laplace filter and then the variance of activations was averaged over 10000 samples images. Higher values are better.}
    \label{tab:smootheness}
\end{table}
\section{Conclusion}
In this paper, we have proposed a new class of Auto Encoders, which we call Pseudo Invertible Encoder. 
We provided a theory that bridges the gap between Auto Encoders and Normalizing Flows. The experiments demonstrate that the proposed model learns the manifold structure and generates sharp images.

\bibliography{references}
\bibliographystyle{iclr2019_conference}

\appendix
\clearpage
\section{Appendix}

\begin{figure}[h!]
\centering
\begin{subfigure}{\linewidth}
    \centering
\includegraphics[width=.3\textwidth]{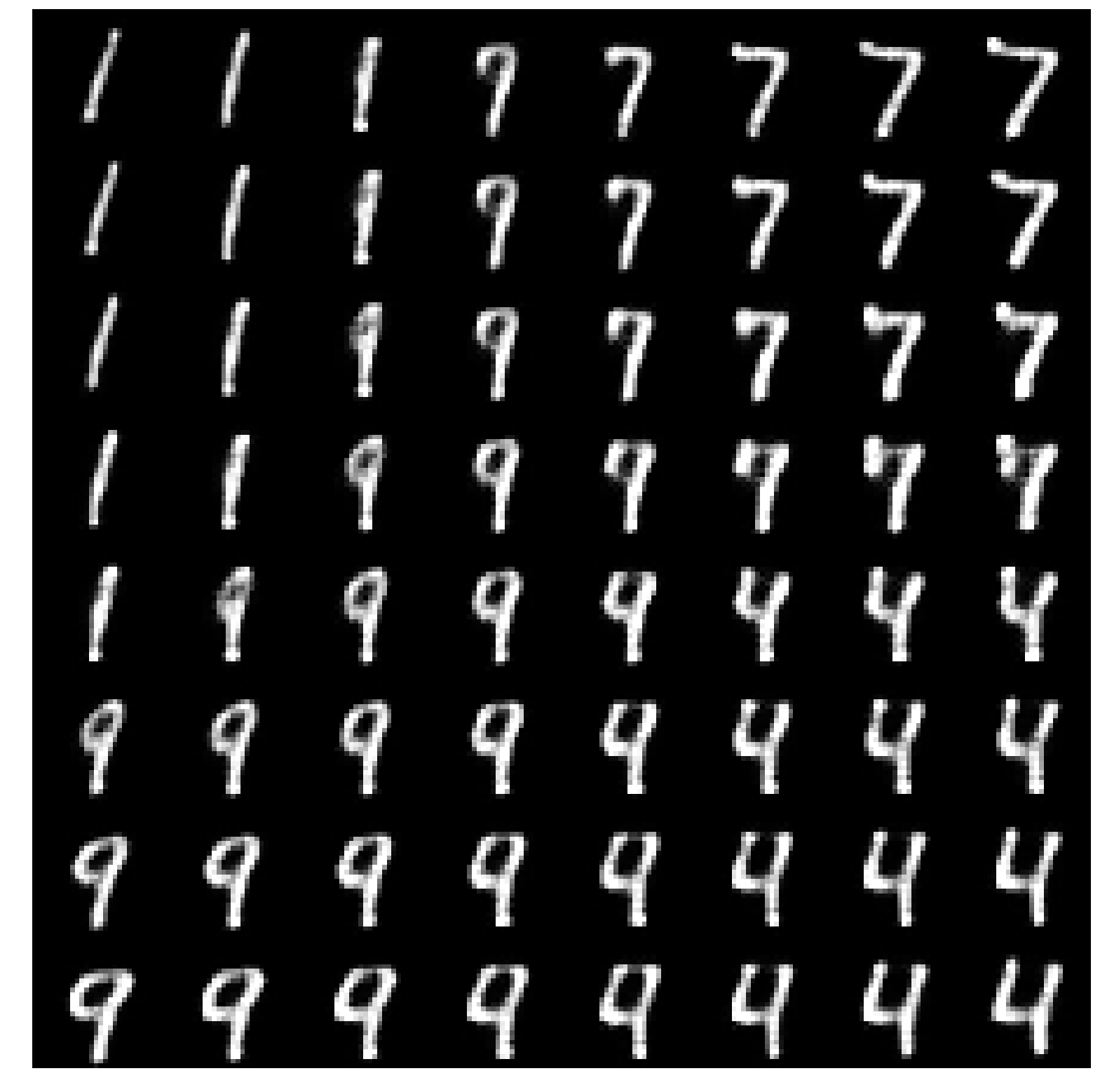}
\includegraphics[width=.3\textwidth]{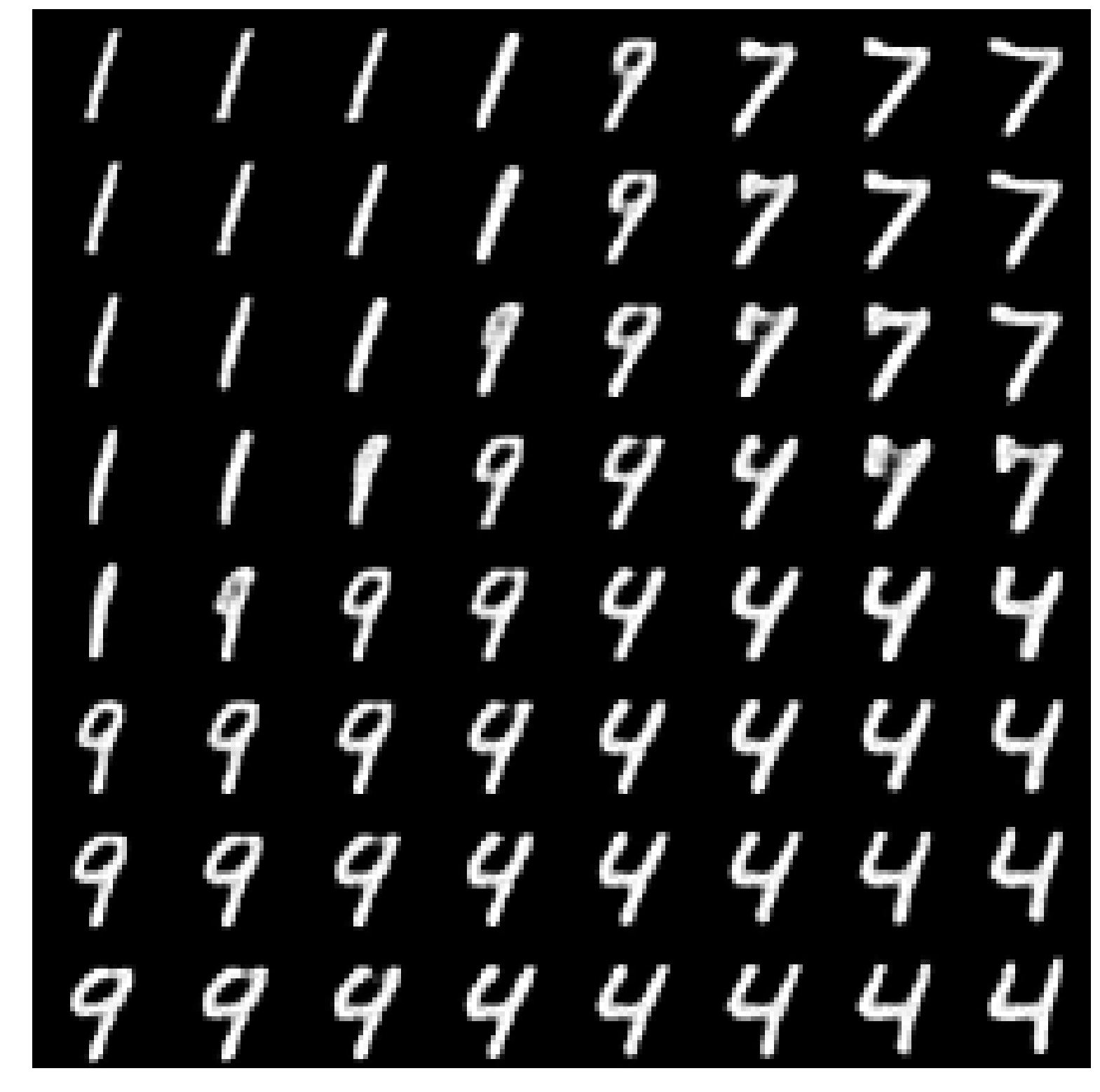} 
\includegraphics[width=.3\textwidth]{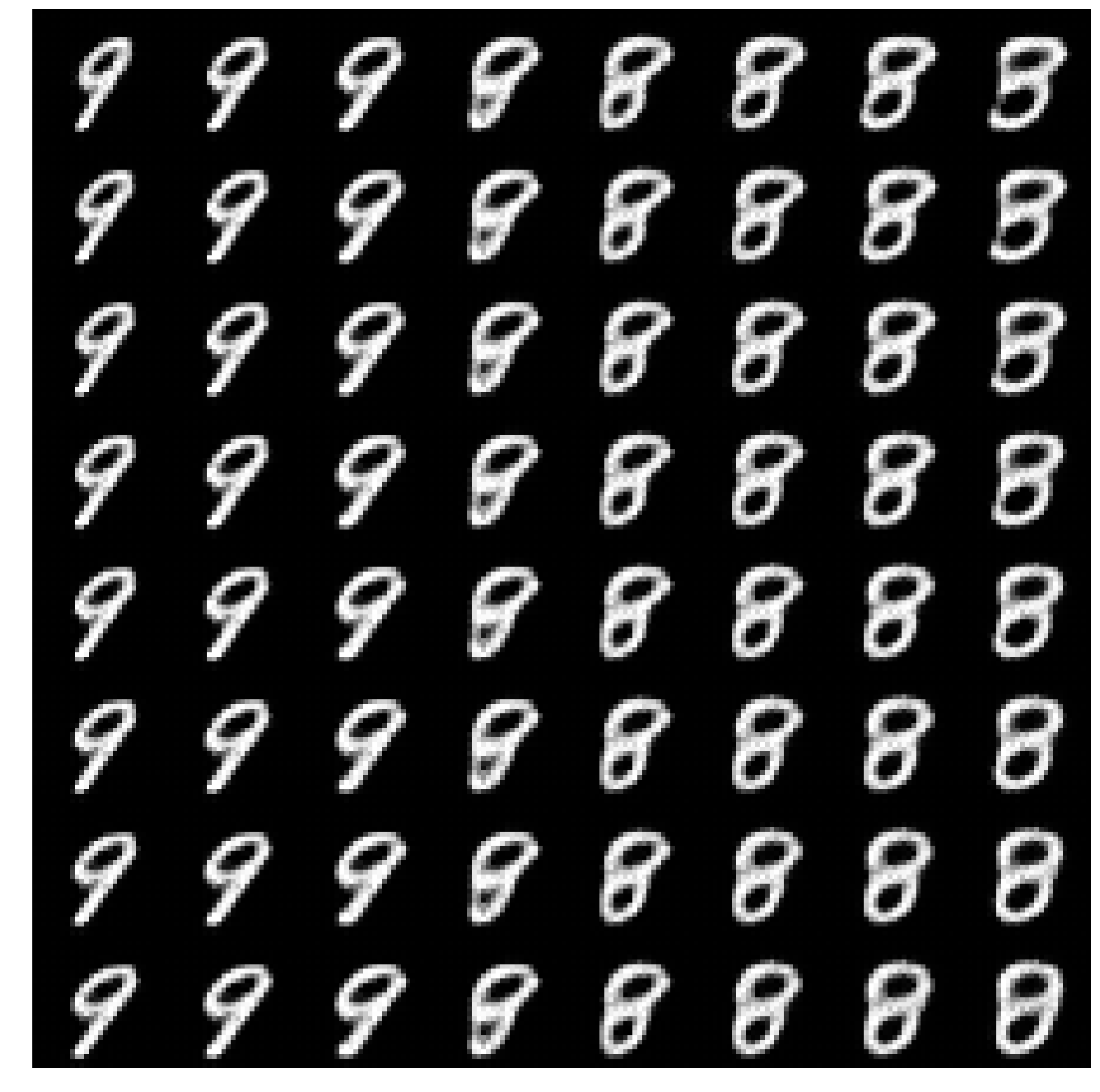}
\end{subfigure}
\begin{tabular}{p{4.2cm} p{4.2cm} p{1.6cm}}
    $\epsilon ^2 = 0.01$ & $\epsilon ^2 = 0.1$ & $\epsilon ^2 = 1.0$ \\ 
\end{tabular}
\caption{Linear interpolation between a picture on the left and the right of the image.}
\end{figure}

\begin{figure}[h!]
\centering
\includegraphics[width=0.9\textwidth]{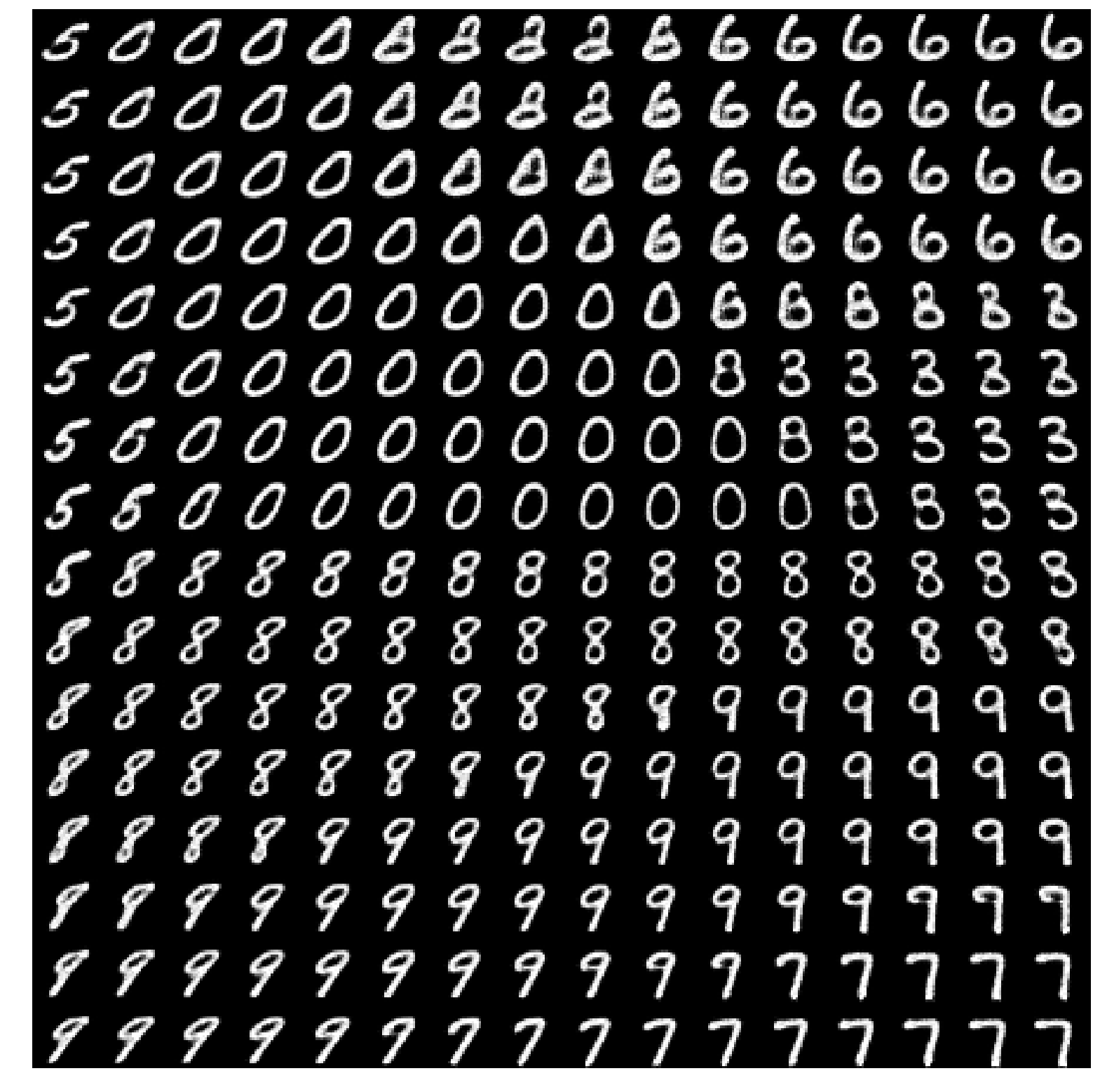}
\caption{Samples from the manifold learned by the Gaussian \textit{PIE}}
\end{figure}
\end{document}